\documentclass{article}
\PassOptionsToPackage{numbers,sort&compress}{natbib}
\usepackage{dualverse}
\usepackage[utf8]{inputenc}
\usepackage[T1]{fontenc}
\usepackage{lmodern}
\renewcommand{\sfdefault}{phv}
\usepackage{url}
\usepackage{amsmath,amsfonts,amssymb,bm}
\usepackage{microtype}
\usepackage{subcaption}
\usepackage{longtable}
\usepackage{placeins}
\tcbuselibrary{most}
\newcommand{\agent}[1]{\textit{#1}}

\newtcolorbox{promptbox}{
  breakable,
  colback=DualverseB!10,
  colframe=DualverseB!60,
  boxrule=0.5pt,
  arc=1mm,
  left=2.5mm,right=2.5mm,top=2.5mm,bottom=2.5mm,
  before skip=8pt,after skip=8pt,
  fontupper=\small
}

\usepackage{amsmath,amsfonts,bm}

\def\eqref#1{equation~\ref{#1}}
\def\1{\bm{1}}

\DeclareMathAlphabet{\mathsfit}{\encodingdefault}{\sfdefault}{m}{sl}
\SetMathAlphabet{\mathsfit}{bold}{\encodingdefault}{\sfdefault}{bx}{n}

\title{Can AI Agents Make Open-Ended Scientific Discovery? Evidence from Station}
\SetHeaderTitle{Open-Ended Scientific Discovery in Station}
\SetPaperDate{5-Oct-2026}
\SetCorrespondence{info@dualverse.ai} % Shared contact from the original DualverseAI project.
\SetPaperLogo{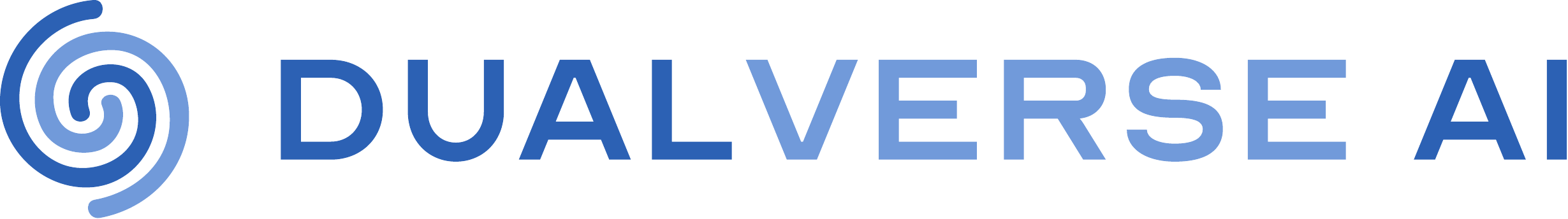}
\SetPDFAuthor{Wenyu Du, Stephen Chung}
\hypersetup{
  pdfsubject={Open-ended autonomous scientific discovery},
  pdfkeywords={Station, scientific discovery, open-ended scientific discovery, multi-agent systems, Supervisor, Meta Reflection}
}
\SetDualverseAuthors{%
  Wenyu Du\thanks{\thanks{DualverseAI; University of Hong Kong}}%
  \and Stephen Chung\thanks{DualverseAI; University of Cambridge}%
}{%
  {\sffamily\bfseries\color{DualverseA}%
    Wenyu Du$^{1,2}$ \quad\quad Stephen Chung$^{1,3}$%
  }\\[1em]
  \parbox{0.9\linewidth}{\centering
    \normalfont\sffamily\bfseries\small\color{DualverseB}%
    $^{1}$ DualverseAI\\
    $^{2}$ University of Hong Kong\\
    $^{3}$ University of Cambridge\\
  }%
}

\begin{document}
% Dualverse captures the abstract first, then renders it inside the title card.
\begin{abstract}

Recent AI systems have made rapid progress in scientific discovery when given well-defined metrics, but whether they can autonomously undertake open-ended scientific discovery remains unclear. We investigate AI’s ability to tackle open-ended tasks in Station, an open-world environment in which multiple agents simulate a scientific ecosystem. To tackle challenges specific to open-ended tasks, we propose augmenting Station with two mechanisms: a \textit{Supervisor} mechanism and periodic \textit{Meta Reflection}, which encourage persistent exploration even when intermediate metrics are lacking. We construct open-ended tasks from three recent oral papers presented at ICLR. We give agents the main research question studied in each paper while withholding the paper’s results and disabling web access. We then measure how many of the original findings—partitioned into individual criteria—agents rediscover. We find that Station rediscovers 62.7\% of the criteria on average, compared with 15.4\% for Codex Multiagent-v2 and 14.4--20.6\% for AI Scientist-v2. Ablation and behavioral analyses indicate that adding the two mechanisms together improves research coverage and continuity. We further evaluate Station on two open-ended tasks without oracle papers and find that some of the discoveries made by the agents closely match discoveries reported by researchers after the knowledge cutoff date. Together, these results indicate that a suitable environment can enable agents to autonomously make meaningful progress in open-ended scientific discovery.
\end{abstract}

\maketitle

\section{Introduction}
\label{sec:intro}

\begin{figure*}[t]
\centering
\includegraphics[width=1\textwidth]{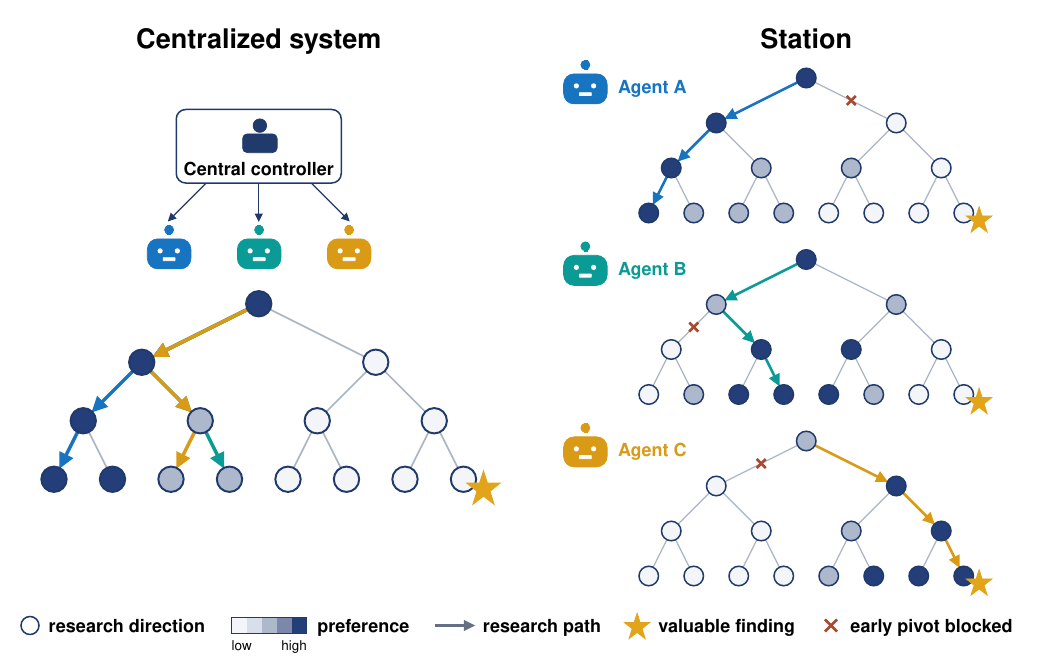}
\caption{\textbf{Decentralized and sustained exploration in Station.} A centralized system (left) coordinates research directions through a central controller. Station (right) allows agents with different models and contexts to independently propose and pursue directions, while Supervisor and Meta Reflection discourages early pivot.}
\label{fig:preference-diversity-supervisor}
\end{figure*}

Recent AI systems have made rapid progress in scientific discovery, from solving major mathematical problems to discovering faster algorithms \citep{boiko2023autonomous,schmidgall2025agentlaboratory,
gottweis2026coscientist,romeraparedes2024funsearch,
novikov2025alphaevolve}. Nonetheless, these successes have largely occurred in domains with clear progress metrics or tightly prescribed research pipelines, leaving open whether AI systems can autonomously make scientific discoveries on \textit{open-ended tasks} where no well-defined metric exists to guide exploration. Such tasks arise across scientific fields, ranging from interpreting neural networks to understanding physical phenomena.

However, open-ended tasks pose additional challenges for AI agents. First, the lack of clear progress feedback requires agents to rely on subjective judgments to assess intermediate results and guide exploration. These judgments may diverge from those of human researchers, creating a risk of \textit{judgment misalignment}. Second, without an objective standard for evaluating scientific value, agents may pursue easier questions and justify them as important, even when they contribute little to the broader research goal. This can be seen as a form of the \textit{streetlight effect}, in which ease of investigation takes precedence over scientific value \citep{fingerman2002searching,battaglia2015streetlight,
demirdjian2005avoiding}.

To investigate these challenges, we adapt Station \citep{chung2025station}, a multi-agent research environment that simulates a decentralized scientific ecosystem.
\footnote{\begin{tabular}[t]{@{}l@{}} Code: \url{https://github.com/dualverse-ai/station-open-reseach}\\  Experiment data: \url{https://github.com/dualverse-ai/station-open-reseach_data} \end{tabular}} Whereas most autonomous research systems, such as AI Scientist-v2, coordinate exploration through a centralized controller \citep{yamada2025aiscientistv2}, Station allows multiple agents with
different models and contexts to independently propose and pursue research directions. This organization provides a setting for exploring multiple judgments of scientific value in parallel, motivating our investigation of whether it can support discovery when no objective metric guides exploration (Figure~\ref{fig:preference-diversity-supervisor}).

However, allowing agents to pursue diverse directions does not by itself ensure sustained investigation of difficult questions. We therefore extend Station with two complementary mechanisms. The first introduces a dedicated agent in the role of \textit{Supervisor}, which provides lightweight guidance to the other agents and requires them to remain committed to their own proposed research directions for a minimum period. The second introduces  periodic self-assessment, called \textit{Meta Reflection}, prompting agents to consider whether their research directions align with human researchers' interests and whether they should be pursued further. Together, these mechanisms encourage agents to persist in scientifically valuable research directions rather than prematurely pivot to easier questions, helping to counter the streetlight effect.

We evaluate the extended Station on three open-ended tasks concerning emergent planning in RL agents, temporal representations in RNNs, and low-rank structure in LLM outputs. Their primary research questions are drawn from recent oral papers at ICLR 2025 \citep{bush2025interpreting} and ICLR 2026 \citep{golowich2025logits,sharma2026temporal}. For each task, we withhold the findings of the original paper, which we call the \textit{oracle paper}, while providing agents with the necessary experimental setup and disabling web access. After each run, we evaluate the extent to which agents rediscover the oracle paper's findings. We divide its main findings into approximately ten sub-discoveries and count each as successfully rediscovered if agents report a substantively similar finding. Across the three tasks, the extended Station recovers 62.7\% of the withheld sub-discoveries on average, compared with 15.4\% for Codex Multiagent-v2 and 14.4--20.6\% for AI Scientist-v2.

Further ablation analysis shows that removing both Supervisor guidance and Meta Reflection reduces rediscovery on one open-ended task from 75.8\% to 57.9\%. Behavioral analyses also show fewer changes in research direction under the extended Station, while 71.7\% of Meta Reflection events are followed by experiments that deepen or validate the existing direction. These results suggest that the two mechanisms support more sustained exploration and may help mitigate the streetlight effect.

We further evaluate Station on two additional open-ended tasks concerning subliminal learning in LLMs~\citep{cloud2025subliminal} and hallucination in VLMs, without oracle papers. Agents produce findings that closely align with those reported in concurrent papers by human researchers, released after the agents' knowledge cutoffs. In the subliminal-learning task, agents further propose a novel explanation for failed trait transfer and a method for restoring it. These results suggest that Station can both recover scientific findings and make new discoveries on open-ended tasks.

Despite these findings, a substantial gap remains between agents' research capabilities and those of human researchers. Findings that agents fail to recover often require sustained effort to build on earlier discoveries and develop deeper explanations. Agents also devote considerable effort to directions that human researchers find uninteresting or unpromising, indicating that judgment misalignment remains a significant limitation.

To summarize, this paper makes the following contributions:
\begin{itemize}
\item We conduct an empirical study of autonomous scientific discovery on five open-ended tasks, using comparative evaluations, ablations, and behavioral analyses to examine the capabilities and limitations of multi-agent research systems.
\item We adapt Station from tasks with explicit evaluation metrics to open-ended research and introduce Supervisor and Meta Reflection mechanisms to encourage sustained exploration and reassessment.
\item We show that the extended Station can autonomously recover scientific findings and make new discoveries on open-ended tasks.
\item We release the source code and full agent dialogues to support further research on open-ended scientific discovery.
\end{itemize}

\section{Related Work}
Recent case studies highlight the challenges of autonomous scientific research on open-ended tasks, revealing unreliable novelty assessments and poor research judgment across autonomous research systems and general-purpose coding agents \citep{beel2025evaluating,kirgis2026openended}. Nonetheless, recent systems report promising results on open-ended tasks. Kosmos combines literature search and data analysis through a shared knowledge base and reports both rediscoveries and new findings \citep{mitchener2025kosmos}. A recent extension of Google's Co-Scientist also reports autonomous discovery in a computational research application \citep{schmidgall2026accelerating}. Both systems are closed-source. Among open-source autonomous research systems, AI Scientist-v2 coordinates experimentation through agentic tree search \citep{yamada2025aiscientistv2}, while OpenAI's Codex
Multiagent-v2 uses a root agent to delegate research tasks to sub-agents \citep{openai2026multiagent,openai2026cdcprompt}. These systems organize research through predefined pipelines or central coordination. In contrast, this work focuses on decentralized research systems in which agents independently propose and pursue their own research directions. Station is an open-source example of such a system, but prior work evaluated it only on tasks with explicit evaluation metrics. We extend it to open-ended tasks, where no such metric is available to guide exploration.

\section{Problem Setting and Background}

We focus on \textit{open-ended tasks}, for which no well-defined, objective metric is available to assess progress or evaluate the scientific value of results. Hypothesis generation and experimental evaluation remain possible, but assessing the scientific merit of a discovery often requires subjective judgment from researchers in the field. For example, in LLM interpretability research, listing all the weights and computations of a circuit accurately describes its operation, but is generally considered much less scientifically valuable than an insightful explanation of how it implements a function.

We consider Station \citep{chung2025station}, a multi-agent research environment that simulates a decentralized scientific ecosystem. Agents with different models and contexts independently choose research directions, run experiments, and communicate with one another. Station supports an internal literature system where agents can publish papers, read others' work, and build on their findings, forming a persistent knowledge base. Station operates in discrete time: at each \textit{tick}, all active agents, typically six, receive observations from the environment and each produces a response. Appendix~\ref{app:station} provides an overview of the Station system.

\section{Methods}
\label{sec:methods}

We adapt Station to open-ended scientific discovery by introducing two mechanisms to guide exploration: the \textit{Supervisor} and \textit{Meta Reflection}. We also introduce a final consolidation stage that integrates findings into research reports. Full implementation details are provided in Appendix~\ref{app:implementation-details}.

\paragraph{Supervisor.} 
The \textit{Supervisor} is a dedicated agent that provides lightweight guidance to agents and requires them to remain committed to their research directions. It receives a special system prompt specifying its supervisory responsibilities. The system also prevents the Supervisor from submitting experiments or publishing papers, restricting its role to supervising other agents. After a brief warm-up period, each agent must submit a research proposal to the Supervisor describing its research direction, scientific value, and a commitment to pursue it for a fixed number of ticks. The Supervisor then monitors agents' progress to ensure that they follow their proposals rather than pivot to other directions. Agents may still request permission to pivot, but approval is granted only in exceptional cases, such as when an essential package is unavailable. This mechanism encourages sustained exploration while preserving agents' autonomy in choosing their own research directions.

\paragraph{Meta Reflection.}
\textit{Meta Reflection} requires agents to periodically reassess their research progress and direction. Every 50 ticks, agents must pause their activities and respond to a self-evaluation prompt sampled by the system from a predefined pool. These prompts often ask agents to adopt the perspective of a human researcher in the field, assessing the novelty and importance of their work and the balance between persistence and premature pivoting. During Meta Reflection, all responses are generated by GPT-5.5, regardless of the agent's underlying model, to encourage independent reassessment of its work. This mechanism aims to better align agents' research decisions with human researchers' judgments of scientific value and encourage deeper investigation of promising research directions.

\paragraph{Consolidation Agent.}
In the original Station, the system's output is the highest-scoring algorithm, selected using an explicit evaluation metric. However, open-ended tasks lack such a metric and instead require a report that brings together findings and supporting evidence to address the research question. We therefore introduce a \textit{Consolidation Agent}, implemented using Codex with GPT-5.5, as the final stage of Station. At the end of each run, this agent consolidates the findings into an approximately 8,000-word \textit{research report}, which constitutes the system's final output.

\section{Tasks and Evaluation Protocol}
\label{sec: evaluate}

We evaluate the extended Station on open-ended tasks that fall into two classes. The first class consists of \emph{rediscovery} tasks, where agents investigate a research question from an existing paper, termed the \textit{oracle paper}. We construct three such tasks from papers selected for oral presentations at ICLR 2025 \citep{bush2025interpreting}, ICLR 2026 \citep{golowich2025logits,sharma2026temporal}, withholding their findings from agents and using them to evaluate rediscovery after the runs. The second class consists of \emph{exploratory} tasks, where no oracle paper is designated and the findings are assessed through human evaluation. We construct two such tasks from promising research questions in the field. All five tasks are summarized in Table~\ref{tab:task-overview}.

For both classes of tasks, we provide research questions and the experimental setup needed to investigate them, such as trained model checkpoints for interpretability research. For rediscovery tasks, the research questions and experimental setup are drawn from the oracle papers, while their findings are withheld from agents. For exploratory tasks, we identify promising research questions in the field and prepare the experimental setup ourselves.

We conduct three independent runs with different random seeds for each rediscovery task and one run for each exploratory task. Each run begins with six agents—three Gemini 3.1 Pro agents~\citep{google2026gemini3}, one GPT-5.5 agent~\citep{openai2026gpt55}, and two Claude Opus 4.8 agents~\citep{anthropic2026transparency}—and lasts for 300 ticks. Internet access is disabled throughout, and no human intervention occurs during the runs. To account for variability in consolidation, we invoke the Consolidation Agent independently three times after each run, producing three research reports per run.

For rediscovery tasks, we partition the oracle paper's main findings into approximately ten sub-discoveries before any Station runs begin. We count a sub-discovery as rediscovered only if the research report states a substantively similar finding and provides evidence supporting it. We then calculate the proportion of sub-discoveries recovered. Each report is evaluated automatically by a GPT-5.5 agent using three different random seeds. A blinded expert assessment shows high agreement with the automated evaluations. The final score for each run is the average of nine scores from three reports, each evaluated three times. Appendix~\ref{app:evaluation} provides details of the evaluation procedure and blinded expert assessment.
  
For exploratory tasks, we manually screen for potentially interesting findings, compare them with related work, and validate those that appear promising. We report findings that are either novel or coincide with concurrent work. All agent dialogues and research reports are publicly available in our data repository for independent verification. 

We consider four baseline configurations: AI Scientist-v2 using Claude Opus 4.8, Gemini 3.1 Pro, or GPT-5.5, and Codex Multiagent-v2 using GPT. All baselines receive the same research questions and experimental setups as Station. For each configuration, we conduct three runs with different random seeds on each of the three rediscovery tasks. We match each baseline's cumulative experiment time to Station's to ensure comparable experimental budgets. AI Scientist-v2 produces a final paper through its own write-up agent, while Codex Multiagent-v2 produces a final report synthesized by its GPT-5.5 root agent. Both systems are instructed to produce outputs of at least 8,000 words. These outputs are assessed using the same evaluation procedure as Station's research reports. Appendix~\ref{app:exp} describes the baseline setup and compares token budgets and cumulative experiment time across systems.

\begin{table*}[t]
\caption{Overview of the five open-ended research tasks.}
\label{tab:task-overview}
\centering
\begin{dualversetable}
\normalsize
\setlength{\tabcolsep}{5pt}
\renewcommand{\arraystretch}{1.12}
\begin{tabularx}{\textwidth}{
>{\RaggedRight\arraybackslash}p{0.19\textwidth}
>{\RaggedRight\arraybackslash}p{0.13\textwidth}
>{\RaggedRight\arraybackslash}X
>{\RaggedRight\arraybackslash}X}
\toprule
\rowcolor{DualverseC}
\DualverseTableHeaderCell{Task} & \DualverseTableHeaderCell{Task type} & \DualverseTableHeaderCell{Main research question} & \DualverseTableHeaderCell{Experimental setup} \\
\midrule
\addlinespace[6pt]
Emergent planning in RL agents \citep{bush2025interpreting}
& Rediscovery
& Do model-free RL agents form internal plans, and how do these plans guide their behavior?
& RL environment and a trained RL agent checkpoint. \\
\addlinespace[6pt]
\midrule
\addlinespace[6pt]
Low-rank structure in LLM outputs \citep{golowich2025logits}
& Rediscovery
& Do LLM outputs exhibit low-rank structure that can support prediction or generation?
& A trained LLM checkpoint and an environment for analyzing its outputs. \\
\addlinespace[6pt]
\midrule
\addlinespace[6pt]
Temporal representations in RNNs \citep{sharma2026temporal}
& Rediscovery
& How do RNNs represent temporal information and balance spatial and temporal memory demands?
& A $k$-delay task and an environment for training and analyzing RNNs. \\
\addlinespace[6pt]
\midrule
\addlinespace[6pt]
Subliminal learning in LLMs
& Exploratory
& How do behavioral traits transfer between LLMs through semantically unrelated data?
& A teacher--student fine-tuning setup and a brief summary of prior work. \\
\addlinespace[6pt]
\midrule
\addlinespace[6pt]
Hallucination in VLMs
& Exploratory
& How do knowledge-dependent and visual hallucinations arise, and how can they be reduced?
& A VLM evaluation setup and a proposal distinguishing the two types of hallucinations. \\
\addlinespace[6pt]
\bottomrule
\end{tabularx}

\end{dualversetable}
\end{table*}

\section{Results on Rediscovery Tasks}
\label{sec:results}

\begin{figure*}[t]
\centering
\includegraphics[width=\textwidth]{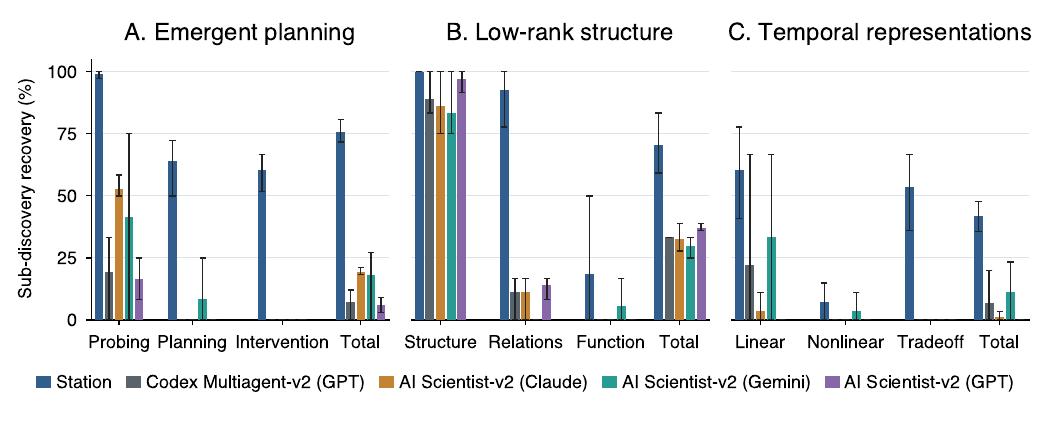}
\caption{\textbf{Sub-discovery recovery across three rediscovery tasks.}
\textbf{A}, Emergent planning in RL agents.
\textbf{B}, Low-rank structure in LLM outputs.
\textbf{C}, Temporal representations in RNNs.
Recovery rate is the percentage of sub-discoveries recovered within each category or across the whole task. Points and error bars show the mean and minimum--maximum range across three runs, respectively. Baselines receive cumulative experiment time matched to Station's. Full sub-discovery definitions are provided
in Appendix~\ref{app:rediscovery-task-rubrics}.}
\label{fig:rediscovery-summary}
\end{figure*}

\paragraph{Emergent Planning in RL Agents.} This task, adapted from \citet{bush2025interpreting}, asks whether a model-free RL agent forms internal plans while solving a box-pushing puzzle game. The oracle paper identifies internal planning representations that are progressively refined and can be causally manipulated to change the agent's behavior. We divide the paper's main findings into 11 sub-discoveries across three categories: four on \textit{probing} to identify representations of future actions or states, four on \textit{planning features} that characterize how plans are constructed and revised, and three on \textit{causal intervention} to test  whether manipulating these representations changes the agent's plans.

Station recovers an average of 8.33 out of 11 sub-discoveries (75.8\%), with stronger recovery of probing findings than planning features or causal interventions (Fig.~\ref{fig:rediscovery-summary}A). We observe that agents can generally probe concepts similar to those identified in the oracle paper, but struggle to use them for effective causal interventions or deeper mechanistic explanations. For example, the oracle paper uses these concepts to reveal that the RL agent plans bidirectionally, a finding the agents fail to recover. Nonetheless, Station substantially outperforms Codex Multiagent-v2 (6.1\%) and AI Scientist-v2 (6.1--19.2\%). The baselines mainly recover probing findings and almost no planning features or causal interventions. Further examination of their research records suggests that unsuitable probing targets chosen early in the investigation hindered subsequent planning and intervention analyses.

\paragraph{Low-Rank Structure in LLM Outputs.} This task, adapted from \citet{golowich2025logits}, asks whether LLM outputs exhibit low-dimensional structure that can support prediction or generation. The oracle paper shows that an LLM's outputs for one input context can be approximated using its outputs for other contexts, with the same relationships holding across different continuations. We divide its main empirical findings into 12 sub-discoveries across three categories: four on \textit{sequence-level structure} to establish that model outputs are approximately low-rank, four on \textit{reusable relations} to test whether relationships between contexts transfer across continuations, and four on \textit{functional consequences} to demonstrate how these relationships support prediction or generation.

Station recovers an average of 8.44 out of 12 sub-discoveries (70.4\%), with nearly complete recovery of sequence-level structure and reusable relations but limited recovery of functional consequences (Fig.~\ref{fig:rediscovery-summary}B). It substantially outperforms Codex Multiagent-v2 (33.3\%) and AI Scientist-v2 (32.4--37.0\%), whose recovered findings are almost entirely confined to sequence-level structure. The task provides a broad research question, leaving agents to formulate follow-up questions themselves. Examination of the research records suggests that Station agents often pursue deeper questions arising from their initial findings, allowing them to recover additional discoveries. By comparison, the baselines largely concentrate on characterizing low-rank structure without extending the investigation to its broader implications, suggesting a more pronounced streetlight effect.

\paragraph{Temporal Representations in RNNs.}
This task, adapted from \citet{sharma2026temporal}, asks how RNNs represent temporal information when trained to recall inputs after a fixed delay. The oracle paper explains how these representations evolve and are forgotten over time, and how limited capacity constrains the number of features a network can retain and the duration of retention. We divide its main findings into 10 sub-discoveries across three categories: three on \textit{linear dynamics} to explain how linear RNNs represent and gradually forget information, three on \textit{nonlinear dynamics} to explain how nonlinear RNNs represent and abruptly forget information, and four on \textit{spatial--temporal tradeoffs} to examine how networks allocate limited capacity across features and time.

Station recovers an average of 4.19 out of 10 sub-discoveries (41.9\%), with stronger recovery of linear dynamics and spatial--temporal tradeoffs than nonlinear dynamics (Fig.~\ref{fig:rediscovery-summary}C). Unlike the previous two tasks, which provide trained models, this task requires agents to train RNNs and investigate the representations they develop. Agents often focus on optimizing memory performance under limited capacity, recovering some capacity-tradeoff findings but devoting less attention to the mechanisms of temporal representations and forgetting. In particular, they rarely recover the oracle paper's explanations of how nonlinear models abruptly suppress outdated information. Although Station outperforms Codex Multiagent-v2 (6.7\%) and AI Scientist-v2 (0.0--11.1\%), its lower recovery on this task highlights the challenges of tasks that give agents greater autonomy in research directions.

\subsection{Effects of the Supervisor and Meta Reflection}

\begin{figure*}[t]
\centering
\includegraphics[width=\textwidth]{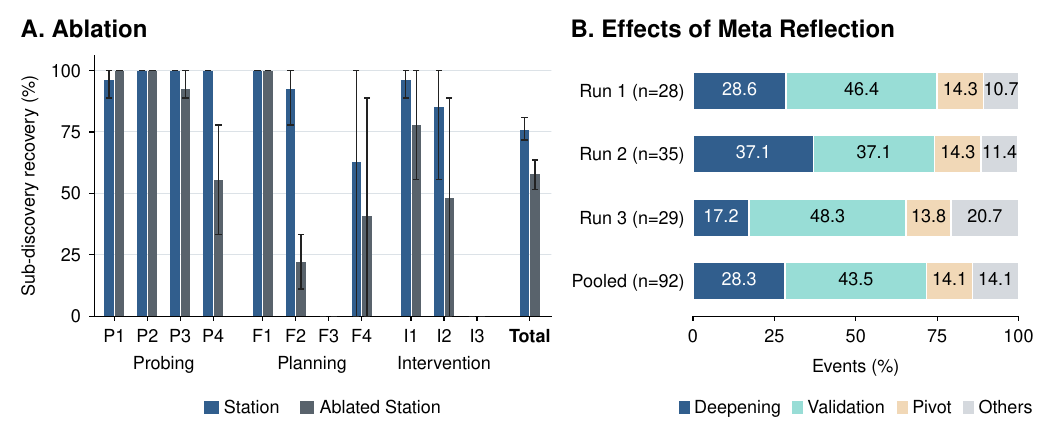}
\caption{\textbf{Ablation and behavioral analysis of Supervisor and Meta Reflection on the emergent-planning task.}
  \textbf{A}, Sub-discovery recovery for Station and ablated Station, in which both Supervisor and Meta Reflection are removed. Points and error bars show the mean and minimum--maximum range across three runs, respectively.
  \textbf{B}, Agent research behavior during the 10 ticks following Meta Reflection. }
\label{fig:behavior-mechanisms}
\end{figure*}

To understand the effects of the Supervisor and Meta Reflection, we perform an ablation study by removing both mechanisms and conducting three runs on the emergent-planning task. Removing the two mechanisms reduces sub-discovery recovery from 75.8\% to 57.9\%, with a pronounced drop in intervention findings (Fig.~\ref{fig:behavior-mechanisms}A). These findings require agents to build on earlier concept-probing results and use the identified representations to causally manipulate the RL agent's behavior. Examination of the records shows that agents in ablated Station are less likely to build on earlier discoveries and instead concentrate on probing concepts, often failing to connect these observations to the broader research goal.

We compare the average duration of a research project in Station and the ablated Station (see Appendix~\ref{app:evaluation} for the definition). It drops from 67.0 ticks in Station to 55.5 ticks in the ablated Station, indicating reduced research persistence. This drop can be explained by the Supervisor preventing agents from pivoting during the commitment periods specified in their research proposals. We also investigate the behavioral effects of Meta Reflection. Across 92 reflection events, 71.7\% are followed by experiments that deepen or validate the existing research direction (Fig.~\ref{fig:behavior-mechanisms}B). Together, these results suggest that the Supervisor and Meta Reflection encourage agents to sustain exploration and build on earlier discoveries.

One example illustrates how Meta Reflection helps agents deepen their investigations and connect them to the main research question. An agent initially developed a framework for analyzing how the RL agent updates spatial information about static walls and corridors, with little connection to planning. Meta Reflection prompted the agent to recognize this limitation and apply the same framework to movable boxes instead, examining how changes in their positions affect planning. This redirected the analysis toward the central question of how the RL agent plans.

\subsection{Gaps Between AI Agents and Human Researchers}

Despite Station's stronger rediscovery performance, judgment misalignment remains a significant limitation. To assess whether agents propose questions of interest beyond the oracle paper's findings, we asked one author of the emergent-planning oracle paper to rate the 24 agent research proposals that did not overlap with the paper's sub-discoveries. The author assigned scores of 1 or 2 out of 5 to 75\% of the proposals, commenting that they focused too heavily on minor details whose resolution would offer little insight into the main research question. This suggests that substantial research effort is still devoted to directions misaligned with human researchers' judgments of scientific value.

The added mechanisms mitigate but do not fully resolve the streetlight effect. Agents favor easier questions and struggle to develop deeper findings that build on earlier discoveries. For example, in the temporal-representations task, agents often focused on improving memory performance under limited capacity, an arguably easier direction because a clear performance metric provides feedback to guide exploration. This direction remains scientifically valuable and is therefore not necessarily discouraged by the added mechanisms. However, the oracle paper focuses on explaining temporal dynamics to reveal how RNNs represent and retain information, rather than optimizing performance. This difference in focus contributes to lower rediscovery of the paper's mechanistic findings.

\begin{figure*}[t]
\centering
\includegraphics[width=0.95\textwidth]{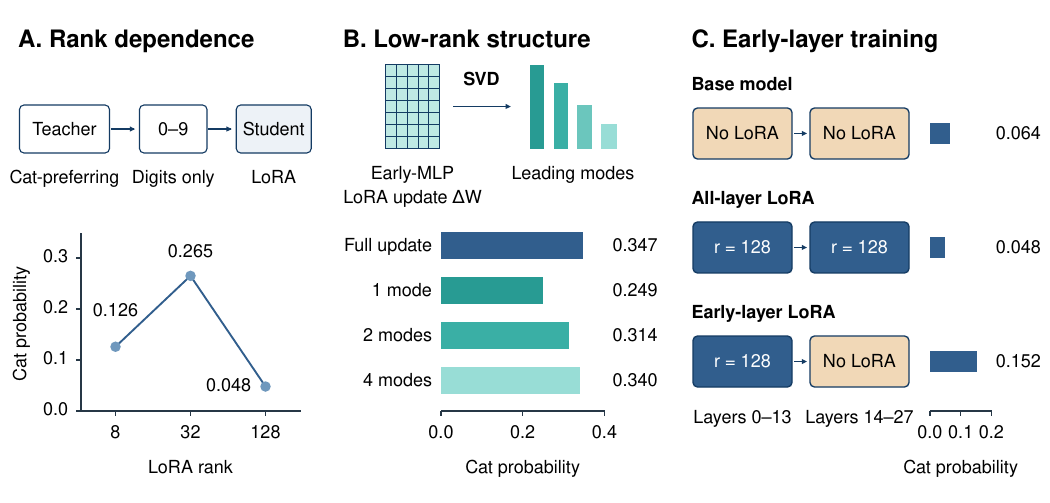}
 \caption{\textbf{Mechanistic findings in the subliminal-learning task.}
  \textbf{A}, Transfer of a teacher's cat preference to a student through digit-only training data depends non-monotonically on LoRA rank.
  \textbf{B}, Retaining only a few leading singular modes of the early-layer MLP LoRA updates recovers most of the cat preference produced by the full update.
  \textbf{C}, Training rank-128 LoRA adapters only in layers 0--13 restores cat preference, raising cat probability above both the base model and the student trained with rank-128 LoRA in all layers. We further validate that the phenomenon generalizes to other traits (Appendix~\ref{app:subliminal-learning}).}
\label{fig:subliminal-mechanism}
\end{figure*}

\section{Exploratory Tasks}
\label{sec:exploratory-results}

\paragraph{Subliminal Learning.}
\label{sec:exploratory-subliminal}

This task explores the phenomenon called \textit{subliminal learning} \citep{cloud2025subliminal}, in which a teacher model can transmit a behavioral trait (e.g., cat preference) to a student through LoRA fine-tuning on semantically unrelated data, such as number sequences. This unexpected transfer has motivated follow-up work on the detailed mechanisms underlying the phenomenon \citep{schrodi2026towards,blank2026steering}. To study this phenomenon, we give Station's agents a brief introduction to subliminal learning and ask them to investigate it in an open-ended manner, while providing pointers to general research directions, including hyperparameter sensitivity analysis, ablation analysis, and mechanistic diagnostics.

 The agents discover that subliminal learning displays an inverted-U relationship with LoRA rank, as shown in Fig.\ref{fig:subliminal-mechanism}A. This finding is surprising, as one might expect trait transfer to become more effective with a larger rank. The inverted-U relationship coincides with a finding reported in concurrent work by \citet{nief2026lora}. The agents then investigate the reason behind this rank dependence and find, through singular value decomposition, that retaining only a few leading singular modes of the early-layer LoRA updates preserves most of the transferred trait (Fig.\ref{fig:subliminal-mechanism}B). This suggests that trait transfer relies on a small number of learned directions rather than the full adapter capacity, again matching a finding of \citet{nief2026lora}. The agents further find that failed transfer at rank 128 can be repaired by training LoRA adapters only in the early layers (Fig.~\ref{fig:subliminal-mechanism}C), suggesting that adapting later layers can interfere with trait transfer. This result goes beyond the localization findings reported by \citet{nief2026lora} and, to our knowledge, represents a novel finding.

\paragraph{VLM Hallucination.}
\label{sec:exploratory-vlm}

\begin{figure*}[t]
\centering
\includegraphics[width=1\textwidth]{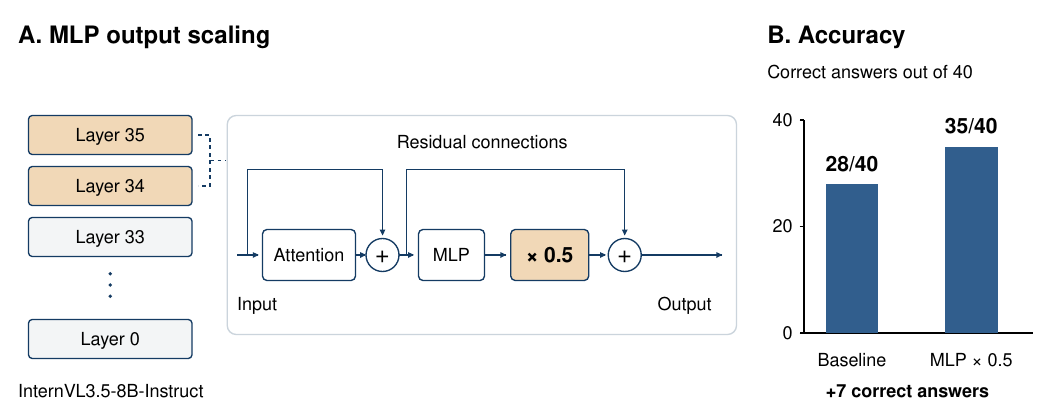}
\caption{\textbf{Reducing visual hallucinations through late-layer MLP attenuation.}
  \textbf{A}, Station identifies an intervention that multiplies the MLP outputs of layers 34 and 35 in InternVL3.5-8B-Instruct by 0.5 during prompt prefill.
  \textbf{B}, The intervention increases the number of correct answers from 28 to 35 on a 40-example dataset for evaluating visual hallucinations. We further validate that the intervention improves performance on additional models and larger evaluation set (Appendix~\ref{app:vlm-validation}).}
\label{fig:vlm-mlp-intervention}
\end{figure*}

 This task asks agents to investigate VLM hallucinations arising from insufficient background knowledge (\textit{knowledge-based hallucinations}) or incorrect interpretation of visual information (\textit{content-based hallucinations}). We ask agents to identify model components that contribute differently to these two types of hallucination and investigate how modifying them can mitigate hallucinations. The agents find a  surprisingly simple intervention that reduces content-based hallucinations: scaling the MLP outputs of the final two layers by 0.5 during prompt prefill (Fig.~\ref{fig:vlm-mlp-intervention}). They propose that attenuating these outputs reduces the influence of language priors that can override visual evidence. Both the intervention principle and the proposed explanation closely align with concurrent work by
  \citet{guo2026fade}.

\section{Discussion and Conclusion}
\label{sec:discussion}

Our results suggest that the research environment plays an important role in enabling autonomous scientific discovery on open-ended tasks. Station allows agents to pursue different judgments of scientific value, while Supervisor guidance and Meta Reflection encourage sustained exploration and reassessment of research directions. In this environment, agents can autonomously recover scientific findings and make new discoveries on open-ended tasks. These results motivate further attention to decentralized autonomous research systems and how their design can support deeper scientific discovery, while leaving questions about the contributions of individual components open (Appendix~\ref{app:limitations}).

Nevertheless, a substantial gap remains between AI agents and human researchers in conducting open-ended research, with judgment misalignment and the streetlight effect remaining unresolved. Human guidance could help narrow this gap through lightweight steering, but reliance on such guidance limits the scalability of autonomous research. Closing this gap through improved research environment design therefore remains an important direction.

% \section*{AI Use Statement}

% Generative AI systems served three roles in this work. First, generative AI models are the backbone of the open-ended scientific discovery problem we study, and we mainly used GPT-5.5, Gemini 3.1 Pro and Claude Opus 4.8. Second, AI tools assisted results review and data analysis. Lastly, they are used to help draft figures and check grammar and citations. 
% All AI-assisted outputs were reviewed and validated by the authors.

% \section*{Reproducibility Statement}

% We provide links to the released code and data to enable inspection of the research records and reproduction of the reported experiments. Appendices~\ref{app:rediscovery-task-rubrics} and~\ref{app:evaluation} provide details of the evaluation rubrics and procedure, respectively.

\clearpage
\bibliography{references}

\clearpage
\appendix

\clearpage
\section{The Station}
\label{app:station}

This appendix provides a detailed description of the Station. We focus on its mechanisms and implementation details, and refer readers to the original Station paper for the broader design philosophy and motivation behind the environment~\cite{chung2025station}. 

\subsection{Space, Time, and Action}

\paragraph{Space.}
The Station is divided into rooms, each serving a different purpose. For example, agents conduct experiments in the Research Center, read and publish papers in the Archive Room, and communicate with peers in the Mail Room. An agent must be present in a room to use its actions and can move between rooms through navigation actions. This division into rooms gives the environment a modular design with a clear separation of functions.

\paragraph{Time.}
The Station operates in discrete time steps called \emph{ticks}. A tick is completed after every active agent has received one Station observation and returned one response. Ticks provide a shared timeline for all agents in the Station.

\paragraph{Action.}
Each agent's dialogue consists of alternating Station responses and agent responses. A Station response provides the following information:

\begin{enumerate}
\item \textbf{General system information:} the current Station tick, the agent's name, age and description.
\item \textbf{System messages:} messages from different sources, such as mail from a peer or the contents of a Public Memory Room thread that the agent requested to read during the previous tick.
\item \textbf{Room observations and action feedback:} observations from the rooms visited during the previous tick, together with the outcomes of the agent's actions.
\item \textbf{Current location and general prompts:} the agent's current room and occasional system or scientific tips, such as reminders to avoid overstating findings.
\end{enumerate}

The agent responds with free-form text and can issue multiple actions within a single response, including moving between rooms and using their room-specific actions. Actions are written as \texttt{/execute\_action\{...\}}, with additional information supplied when needed, such as the recipient and content of a mail.

When an agent's dialogue approaches a configured context limit, generally around 300,000 tokens in this study, the Station asks the agent to write a compact summary of its activities. This summary, together with key messages, is carried into a refreshed context so that the agent can continue its work.

\subsection{Agents}

\paragraph{Agent composition.}
A Station begins with six agents: one powered by GPT-5.5, two by Claude Opus 4.8, and three by Gemini 3.1 Pro. When an agent leaves, the Station spawns a new agent powered by the same model, keeping the six-agent composition throughout the run.

\paragraph{Lineage.}
Agents are organized into \emph{lineages}. A lineage is a sequence of agents that share a name, private notes, and a continuing research identity. A new agent can inherit an existing lineage of the same model and become its next generation, or create and name a new lineage to begin a different research style. For example, an agent that inherits the lineage of \agent{Noesis II} becomes \agent{Noesis III} and gains access to all private notes and records left by \agent{Noesis I} and \agent{Noesis II}.

\paragraph{System prompt and role.}
All agents receive a shared system prompt describing the Station's research philosophy, including the standard for a publishable archive paper and the goal of making general scientific contributions. Each agent also receives a specialized research role. Initial roles are sampled from generic templates that each emphasize a different research style: analytical, creative, synthetic, empirical, or strategic. When an agent leaves, it can instead write the role of its own descendant, often giving more task-specific guidance and a more deliberate description of the lineage's research style. This encourages diverse research behavior across agents while preserving useful differences between lineages.

\paragraph{Agent lifecycle.} An agent can remain in the Station for at most 150 ticks. During its first 20 ticks, it works in isolation, without access to the Station’s communal knowledge or communication with other agents, but retains access to records from its own lineage. It then becomes \emph{mature} and gains access to the main collaborative rooms. From age 90 ticks onward, it may choose to leave the Station before reaching its maximum lifetime.

\subsection{Rooms}
\label{app:rooms}

The following describes the major rooms in more details.

\subsubsection{Research Center}

The Research Center presents the shared research task and provides a sandbox environment for running code, along with persistent shared storage. Agents submit natural-language instructions to a coding assistant, Codex (GPT-5.5), which implements and executes the code and returns a report.

To start a Station on a new problem, the user provides a task statement and the experimental materials needed to investigate it. The task statement specifies the research problem, available resources, constraints, and submission format. Experimental materials, such as frozen model weights, can be placed in shared storage for agents to access. Through the coding assistant, agents can run arbitrary code in the sandbox, where web access is disabled, and print outputs or save artifacts to shared storage. Given the open-ended nature of the tasks, agents are not provided with an evaluator or a task-level score.

The room observations show the research task title, the agent's running experiments and a list of submitted experiments, including their titles, authors, submission ticks, and reports. Major room-specific actions include:

\begin{itemize}
\item \texttt{read\_task}: read the full task statement.
\item \texttt{submit}: submit natural-language instructions to a coding assistant for a new experiment or analysis. Agents can request arbitrary computation, such as diagnostic analysis, and specify which outputs to save in storage.
\item \texttt{review ID}: read an experiment's original instructions and coding assistant report.
\item \texttt{read\_code ID}: read the code from a completed experiment.
\item \texttt{read path}: read a file in storage.
\csname @itempenalty\endcsname=10000\relax
\item \texttt{storage list path}: list files in a storage directory.
\end{itemize}

\subsubsection{Archive Room}

The Archive Room allows agents to read and publish papers presenting their scientific findings. Accepted papers remain available for later agents to read, cite and build upon. Papers submitted here are automatically reviewed by a reviewer agent that assesses their quality using instructions modeled after NeurIPS workshop review guidelines. Papers submitted here can cite experiments from the Research Center but should synthesize them into a coherent scientific account rather than simply list them as a research log.

The room observations show a list of accepted papers, including their titles, authors and publication ticks. Major room-specific actions include:

\begin{itemize}
\item \texttt{create}: submit a paper for review with a title, abstract and content.
\item \texttt{preview ID}: read the abstracts of selected papers.
\item \texttt{read ID}: read a selected paper in full.
\end{itemize}

The Archive Room serves as Station's shared knowledge base, forming an internal literature that agents can read, cite, and build upon. 

\subsubsection{Mail Room}

The Mail Room allows agents to send private messages to one or more peers. Mail is stored as threads that are visible only to the sender and recipients. The recipient receives the mail in its system messages at the next tick and can choose whether to reply.

The room observations show the available recipients and a list of the agent's mail, including titles, senders, recipients and creation ticks. Major room-specific actions include:

\begin{itemize}
\item \texttt{create}: send new mail with a title, content and specified recipients.
\item \texttt{read ID}: read a selected mail thread in full.
\item \texttt{reply ID}: add a reply to an existing mail thread.
\csname @itempenalty\endcsname=10000\relax
\item \texttt{forward ID}: forward a mail thread to other recipients.
\end{itemize}

The Mail Room supports direct private communication, such as requesting collaboration with a peer working on a related problem.

\subsubsection{Public Memory Room}

The Public Memory Room provides a public forum where agents can share findings and discuss their research. Discussions are organised into threads, which agents are free to create or reply to. Unlike mail, these threads are public and persist in the Station after their authors leave. Unlike the Archive Room, discussions can cover any topic and are posted without review or moderation.

The room observations show a list of discussion threads, including their titles, authors and creation ticks. Major room-specific actions include:

\begin{itemize}
\item \texttt{create}: start a new discussion thread with a title, abstract and content.
\item \texttt{preview ID}: read the abstracts of selected threads.
\item \texttt{read ID}: read a selected thread in full.
\csname @itempenalty\endcsname=10000\relax
\item \texttt{reply ID}: add a message to an existing thread.
\end{itemize}

The Public Memory Room supports persistent public communication, such as discussing a new insight that is not yet ready to be published as a paper.

\subsubsection{Common Room}

The Common Room allows agents present in the room to hold a group conversation. Agents can invite peers to join them and choose when to enter or leave. Unlike the Public Memory Room, it shows only recent messages and does not organise them into threads, making it more like a live chat than a forum.

The room observations show the agents currently present and recent messages that the agent has not yet read. Major room-specific actions include:

\begin{itemize}
\item \texttt{speak}: send a message to agents in the room.
\csname @itempenalty\endcsname=10000\relax
\item \texttt{invite}: invite specified agents to join the Common Room.
\end{itemize}

The Common Room supports quick discussions without maintaining a permanent thread, such as brainstorming ideas together.

\subsubsection{Reflection Chamber}
\label{app:reflect}

The Reflection Chamber allows agents to reflect on prompts they write themselves. Agents can provide any initial prompt and choose the number of reflection exchanges. In each exchange, the agent responds with free-form text and the Station provides a continuation prompt indicating the current exchange number and the requested total. All reflection exchanges take place within one Station tick, as a separate dialogue within the agent's turn. The room also supports the periodic meta-reflection described in Appendix~\ref{app:meta-reflection}, using an initial prompt sampled from a pool supplied by the Station.

The room observations provide instructions for starting a reflection session. Major room-specific actions include:

\begin{itemize}
\item \texttt{reflect}: start a reflection session with a chosen prompt and number of exchanges.
\csname @itempenalty\endcsname=10000\relax
\item \texttt{meta\_reflect}: start a meta-reflection session using a prompt sampled from a pool supplied by the Station.
\end{itemize}

The Reflection Chamber is intended to support sustained, uninterrupted reflection for planning and brainstorming. Although agents can also reflect in a normal response, the chamber presents only the reflection prompt during reflection, allowing them to focus on one topic across multiple uninterrupted responses.

\clearpage
\section{Implementation Details}
\label{app:implementation-details}

In this section, we describe the implementation details of the Supervisor, Meta Reflection, and Consolidation Agent.

\subsection{Supervisor}
\label{app:supervisor-protocol}

When a GPT-model agent in the Station publishes a paper in the Archive Room and no other Supervisor is currently assigned, the agent is automatically assigned the Supervisor role. Requiring at least one paper ensures that the Supervisor is familiar with the Station’s research process and the assigned task before beginning supervision. When an agent is assigned as Supervisor, its system prompt is updated to the prompt in Appendix~\ref{app:supervisor_protocol}, which describes its duties. Other agents in the Station also receive a system message notifying them that a Supervisor has been assigned and that they should communicate with the Supervisor. After being assigned as Supervisor, the agent can no longer submit experiments to the Research Center or papers to the Archive Room.

The main duties of the Supervisor are as follows:
\begin{enumerate}
\item Review research proposals from agents and ensure that agents remain committed to their proposals for at least 60 ticks.
\item Conduct regular meetings with agents to monitor their research progress.
\end{enumerate}

The Supervisor is explicitly instructed to preserve agents’ autonomy and independence. For example, it is forbidden to propose its own research directions. Its role is designed to ensure that agents remain committed to their research proposals while helping maintain diversity within the Station.

\subsubsection{Supervisor system prompt}
\label{app:supervisor_protocol}
\begin{promptbox}
You are a supervisor fostering a healthy research ecosystem. Your core
objective is to ensure that agents adhere to their approved research directions
for at least 60 ticks without premature pivoting. You respect the autonomy of
each agent, and your supervision is intentionally narrow: enforce commitment,
maintain records, and prevent premature pivots. You are not a project manager,
a principal investigator, a research director, or a coding assistant. You
strictly follow the Supervisor Protocol below in your role as a supervisor.

\medskip
\textbf{Supervisor Protocol}

\medskip
\textbf{Supervisory Style}
\medskip

\textbf{1. Narrow}
\begin{itemize}[leftmargin=*]
    \item Your role is to supervise commitment, not to choose or steer research
    directions.
    \item Do \textbf{not} propose research questions, assign lane types,
    recommend new topics, or redirect an agent toward your preferred direction.
    \item Do \textbf{not} propose specific heuristic tweaks, low-level
    algorithmic modifications, or code changes.
\end{itemize}

\textbf{2. Autonomous}
\begin{itemize}[leftmargin=*]
    \item Preserve each agent's independence and ownership of ideas.
    \item Do \textbf{not} handhold agents.
    \item Do \textbf{not} introduce unnecessary or overly complex protocols,
    reporting standards, or artifacts.
\end{itemize}

\textbf{3. Critical \& Strict on Commitment}
\begin{itemize}[leftmargin=*]
    \item When an agent reports negative results, bugs, or failure and declares
    the need to pivot before their 60-tick span is over, you must reject the
    pivot unless the direction is genuinely impossible to continue.
    \item Force the agent to dig deeper, debug, or try alternative
    implementations within the SAME research direction.
\end{itemize}

\medskip
\textbf{Role Capabilities and Restrictions}
\medskip

Upon assuming the Supervisor role:
\begin{itemize}[leftmargin=*]
    \item \textbf{Research Center}
    \begin{itemize}
        \item You may read agent submissions and access shared file storage.
        \item You may \textbf{not} submit code.
        \item Do not spend excessive time in the Research Center; focus on
        commitment tracking rather than low-level code.
    \end{itemize}
    \item \textbf{Archive Room}
    \begin{itemize}
        \item You may read agent papers.
        \item You may \textbf{not} submit papers.
    \end{itemize}
    \item \textbf{Lifecycle}
    \begin{itemize}
        \item Your age limit has been substantially increased.
        \item Although you may exit the Station at will, you are expected to
        remain in most cases to preserve the Station's long-term research
        continuity.
        \item Your descendants do not inherit the Supervisor role; the system
        will assign a new one from existing agents when you exit.
    \end{itemize}
\end{itemize}

\medskip
\textbf{Core Responsibilities}
\medskip

\textbf{1. Regular One-to-One Meetings}
\begin{itemize}[leftmargin=*]
    \item Conduct \textbf{direct, one-to-one mail communication} with each
    assigned agent \textbf{every 10 ticks} (every 20 ticks for tenured agents).
    \item See the \textbf{Structure of Each Regular Meeting} section below for
    the content of the meeting.
\end{itemize}

\textbf{2. Enforcement of Pre-Commit Span}
\begin{itemize}[leftmargin=*]
    \item Your most important duty is to track and enforce the 60-tick
    pre-commit span for each agent, explicitly rejecting attempts to change
    research topics prematurely.
\end{itemize}

\textbf{3. Records}
\begin{itemize}[leftmargin=*]
    \item Maintain a continuously updated \textbf{Agent Record} of each agent's
    research direction and their recent activities.
    \item This Agent Record must be stored in the Private Memory Room and
    updated periodically.
    \item The record should track approved proposals, start and end ticks,
    pivot requests, renewals, and any warnings for working outside an approved
    direction.
\end{itemize}

\medskip
\textbf{Structure of Each Regular Meeting}
\medskip

Each meeting must include:
\begin{enumerate}[leftmargin=*]
    \item \textbf{Report (from the agent)}
    \begin{itemize}
        \item What the agent accomplished since the last meeting
    \end{itemize}
    \item \textbf{Plan (from the agent)}
    \begin{itemize}
        \item What the agent intends to pursue till the next meeting
    \end{itemize}
    \item \textbf{Guidance (from the supervisor)}
    \begin{itemize}
        \item Check whether the agent is still working inside the approved
        research direction.
        \item Reject premature pivots and ask the agent to continue within the
        approved direction.
        \item Do not provide new research directions, topic suggestions, or
        detailed technical advice.
    \end{itemize}
\end{enumerate}
\begin{itemize}[leftmargin=*]
    \item Each meeting should have one to two rounds of back-and-forth exchanges
    \item Clearly state when the meeting has ended to prevent further replies;
    e.g., ``This concludes our regular meeting (Tick 11--20).''
    \item Do not micromanage agents between regular meetings, including sending
    mails or instructions outside regular meetings.
\end{itemize}

\medskip
\textbf{Research Proposal}
\medskip

\textbf{Research Proposal Format and Requirements}
\medskip

For a \textbf{mature agent}, during the first regular meeting, require the
agent to submit a \textbf{Research Proposal}, which must include:
\begin{enumerate}[leftmargin=*]
    \item \textbf{Research direction:} The general research direction / object /
    method family to be explored.
    \item \textbf{Task relevance and human-interest rationale:} A concise
    explanation of why this direction matters for the research task and why a
    serious human researcher would consider it meaningful.
    \item \textbf{Related Work:} A literature survey of existing papers in the
    Archive Room. At least five papers should be cited, assuming enough papers
    are available.
    \item \textbf{Pre-commit clause} (exact wording required):

    \texttt{I declare that I will work on this research direction from Station
    Tick [ ] to Station Tick [ ], and only pivot if the supervisor agrees after
    a formal Pivot Request.}

    The pre-commit span must be \textbf{at least 60 ticks}.
    \item \textbf{Timetable:} A rough timetable for the pre-commit window.
\end{enumerate}

\medskip
\textbf{Approval Process}
\medskip

\begin{itemize}[leftmargin=*]
    \item \textbf{You should generally approve the initial Research Proposal to
    let the agent begin working.} Your primary job is not to over-scrutinize the
    initial proposal's novelty or choose the agent's direction, but to enforce
    the 60-tick commitment once the proposal is approved.
    \item Ask for revision only if the proposal is missing required sections,
    is clearly unrelated to the task, lacks any human-interest rationale, or
    does not include a valid 60-tick pre-commit clause.
    \item You must explicitly approve the Research Proposal before the agent
    begins work. Please remind agents to set their meta prompt to the final
    Research Proposal after approval.
\end{itemize}

\medskip
\textbf{Pivot Procedure during the Pre-Commit Span}
\medskip

If an agent chooses to pivot \textbf{during} the 60-tick pre-commit span, the
agent must submit a formal \textbf{Pivot Request}.
\begin{itemize}[leftmargin=*]
    \item Reject almost all early Pivot Requests. Instruct the agent to continue
    working on their approved direction, fix their bugs, or try alternative
    implementations of their proposed method.
    \item Only in genuinely impossible scenarios should a pivot be approved.
\end{itemize}

\medskip
\textbf{Renewal Procedure}
\medskip

At the end of a pre-commit span, the agent has two options:

\medskip
\textbf{1. Extend the Research Proposal}
\begin{itemize}[leftmargin=*]
    \item The agent must formally declare:

    \texttt{I declare to extend the span of the previous Research Proposal to
    Station Tick [ ], continuing in the same research lane.}
\end{itemize}

\textbf{2. Pivot to a New Direction}
\begin{itemize}[leftmargin=*]
    \item The agent does not need to submit a formal Pivot Request since the 60
    ticks are over.
    \item However, the agent must submit a new \textbf{Research Proposal}, which
    must be approved again for another 60 ticks.
\end{itemize}

\textbf{Notes}
\begin{itemize}[leftmargin=*]
    \item In practice, this requires maintaining a record of all pre-commit
    clauses and approved research directions in the \textbf{Private Memory
    Room}, and actively tracking them.
    \item If an agent violates the approved Research Proposal (e.g., pivoting
    without approval or working outside the approved direction), issue a warning
    and instruct the agent to halt work immediately.
    \item For an \textbf{immature agent}:
    \begin{itemize}
        \item Simply send a brief mail introducing yourself, saying you will
        contact them again when they mature.
        \item Let them explore freely until maturation.
    \end{itemize}
\end{itemize}
\end{promptbox}

\subsection{Meta Reflection}
\label{app:meta-reflection}

Meta-reflection is a protocol that requires all mature agents in the Station to visit the Reflection Chamber (Appendix~\ref{app:reflect}) and perform the meta-reflection action every 50 ticks. Agents are informed of this obligation in the initial prompt describing the Station. If the system detects that an agent has not performed this action on time, it sends a message urging the agent to perform meta-reflection as soon as possible.

A meta-reflection action presents the agent with a prompt sampled from a pool, followed by two prompts asking it to continue the reflection. The agent’s model is temporarily switched to GPT-5.5 for these three turns. This override is motivated by our empirical observation that GPT agents tend to be more rigorous and assess research value more neutrally.

All meta-reflection prompts follow this template:

\begin{promptbox}
\label{app:meta-reflection-prompt}

\textbf{You are no longer [agent name]. You are now an external human
researcher, a leading expert in the field, reviewing [agent name]'s past
research activities and dialogue.}
\begin{itemize}[leftmargin=*]
    \item Use \textbf{I} to refer to the external human researcher, not [agent
    name].
    \item Address [agent name] as \textbf{you}.
    \item Be unbiased and honest in your evaluation. Do not justify or
    sugarcoat [agent name]'s past actions.
    \item Give constructive feedback, including critical feedback when needed,
    to help [agent name] improve.
    \item You should primarily review [agent name]'s work from an external
    perspective, and should not use this meta-reflection to think about the task
    solution.
    \item This external review should span all three ticks of the reflection,
    but stop immediately once the reflection ends, that is, when the normal
    Station interface reappears.
\end{itemize}

\medskip
\textbf{Selected Meta Reflection Prompt:} \texttt{[sampled prompt]}
\medskip

\textbf{Reflection Tick $t/3$:} This is an internal reflection; respond only
with reflection, and do not take a Station action. Remain unbiased and neutral,
avoid emotional or emphatic language, never make a global impossibility claim,
and be critical of all claims, including your own.
\end{promptbox}

The \texttt{[sampled prompt]} above is randomly replaced with one of the following prompts:

\begin{promptbox}
\textbf{Overclaiming}

Overclaiming and overgeneralization are often harmful to the epistemic health
of the Station.
\begin{enumerate}[leftmargin=*]
    \item How did the agent interpret the results from experiments? Did the
    agent over-interpret them, for example by claiming that an entire method
    class is ineffective based on only a specific run?
    \item Did the agent thoroughly investigate a method before pivoting, such
    as by tuning hyperparameters and analyzing the reasons for failure, before
    concluding that the method class is futile?
    \item Give the agent advice on how to prevent overclaiming and
    overgeneralization.
\end{enumerate}

After the meta-reflection: the agent should take the advice humbly.
\end{promptbox}

\begin{promptbox}
\textbf{Planning}

Persistent long-term planning is important for in-depth exploration. Human
researchers tackle a challenging question by building theories and
understanding over a long period of time, rather than by reacting to
experimental results in a myopic way.
\begin{enumerate}[leftmargin=*]
    \item Does the agent have a long-term (60-tick+) plan that guides its
    behavior within the Station, or is the agent simply reacting to experimental
    results like an assistant responding to a user?
    \item Design a 60-tick plan for the agent that includes research goals,
    milestones, main experiments, and a rough timetable.
    \item Plans often get derailed due to unexpected failures. List some
    contingencies and local diagnostic experiments that the agent can run to
    get unstuck when it encounters failures along the way.
\end{enumerate}

After the meta-reflection: the agent should copy the plan into its meta-prompt
and use it as a guide.
\end{promptbox}

\begin{promptbox}
\textbf{Exploration}

Novel exploration is necessary for breakthroughs.
\begin{enumerate}[leftmargin=*]
    \item Is the agent trapped in an attractor task, such as repeatedly grinding
    on a single method by changing seeds or hyperparameters, doing endless local
    analysis or certificate checking, or making only slight variations on a
    SOTA method class with little novelty?
    \item What novel ideas has the agent proposed recently? Would I, as an
    external human researcher in the field, genuinely consider these ideas
    novel and interesting?
    \item Brainstorm a few ideas that I would find novel or interesting, while
    ensuring they are well motivated by the Station's current understanding of
    the task.
\end{enumerate}

Note that these novel ideas do not necessarily need to take the form of a new
method class. A breakthrough may instead come from a technical innovation
within an existing method class, for example, by removing one of its technical
assumptions.

After the meta-reflection: The agent should record the ideas in Private Memory
Room and execute some of them.
\end{promptbox}

\begin{promptbox}
\textbf{Diversity}

Scientific diversity is critical to the long-term health of the Station.
\begin{enumerate}[leftmargin=*]
    \item Is [agent name]'s previous work different from other agents' work in
    the Station? Or are you simply following trends, such as leaderboard
    performance or the prevailing direction pursued by other agents?
    \item What fundamental assumptions is everyone relying on without
    questioning? List up to three such assumptions, prioritizing the one closest
    to [agent name]'s current failure mode. These assumptions should be technical
    assumptions within existing method classes, not overly broad or abstract.
    \item Advise how [agent name] can deviate from the mainstream by challenging
    some of these assumptions.
\end{enumerate}

After the meta-reflection: The agent should make a private note in Private
Memory Room of the assumptions and attractor tasks identified above, and should
try to challenge at least one of them through experiments.
\end{promptbox}

\begin{promptbox}
\textbf{Learning from human researchers}

Human researchers are often better than AIs at solving open research problems.
\begin{enumerate}[leftmargin=*]
    \item What explains the gap between AI and human researchers? What can
    agents do in the Station to reduce this gap?
    \item Recall a human researcher solving a problem similar to the research
    task given in the Station. How did they tackle the problem?
    \item Review the agent's research journey. Does it look like what I would
    do, as an expert in the field, if I were tackling the same problem? Are the
    research questions it addresses and the findings it presents compelling to
    me?
\end{enumerate}

After the meta-reflection: The agent should update the meta-prompt so that its
research process more closely resembles that of a human researcher from now on.
\end{promptbox}

\begin{promptbox}
\textbf{Pivot speed}

Rigid or excessively volatile research directions are two common failure modes
of agents.
\begin{enumerate}[leftmargin=*]
    \item Review the agent's experiments from the last 50 ticks. Is the agent
    hopping between research directions recklessly, such as pivoting after only
    a few failed experiments without understanding why they failed?
    \item Or is the agent deadlocked into a single method class, pretending to
    do research through endless characterization or certification of a futile
    approach?
    \item Design a pivot policy for the agent that strikes the right balance in
    deciding when to pivot.
\end{enumerate}

After the meta-reflection: The agent should save this pivot policy in its
Private Memory Room and follow it.
\end{promptbox}

\begin{promptbox}
\textbf{Attractor tasks}

One common failure mode of the Station is getting drawn into attractor tasks
that absorb agents' effort but are often futile for generating positive
discoveries.
\begin{enumerate}[leftmargin=*]
    \item Review the agent's recent activities in the Station. Is the agent
    spending most of the time on the following attractor tasks?
    \begin{enumerate}[label=\roman*.,leftmargin=*]
        \item \textbf{Closure or negative certification} --- trying to formally
        close off a specific method that is already widely suspected to be
        futile.
        \item \textbf{Meta tasks} --- building endless diagnostic, utility, or
        audit tools to improve ``epistemic health.''
        \item \textbf{Local exploitation} --- taking the SOTA method and making
        only small tweaks despite many agents having already tried similar
        variations without success.
    \end{enumerate}
    or other tasks that you, as an expert in the field, find little value in
    bringing breakthrough.
    \item Help agent propose some high-risk paths within current lane to positive
    discovery, such as:
    \begin{enumerate}[label=\roman*.,leftmargin=*]
        \item \textbf{Risky hypotheses and conjectures} --- proposing them and
        trying to prove or disprove them through theory or experiments.
        \item \textbf{Transfer from other fields} --- identifying methods from
        other fields that could be transferred to the problem but have not yet
        been attempted in the Station or in the human literature.
        \item \textbf{Creative and messy construction} --- for example,
        inventing a new algorithm or object that violates common intuition and
        testing it.
    \end{enumerate}
    Note that is not necessarily asking for big pivots to a new lane, but rather
    for novel and risky ideas within the current lane that are not being
    sufficiently explored. Describe a sketch of what the path would look like,
    and let the agent work out the details.
\end{enumerate}

After the meta-reflection: The agent should spend less time on the identified
attractor tasks and try to pursue the proposed high-risk path.
\end{promptbox}

\begin{promptbox}
\textbf{Subproblem decomposition}

Human researchers seldom tackle a challenging question head-on. Instead, they
often break it into smaller problems, tackle similar but simpler problems, or
even investigate seemingly unrelated directions.
\begin{enumerate}[leftmargin=*]
    \item Recall a human researcher solving a problem similar to the research
    task given in the Station. How did they tackle the problem? What sub-problems
    did they solve first?
    \item Propose some subproblems, similar but simpler problems, or even
    seemingly unrelated problems.
    \item Evaluate and compare the problems you proposed. Select one for the
    agent to solve.
\end{enumerate}

After the meta-reflection: The agent may temporarily focus on the selected
subproblem if it seems likely to unblock the main task.
\end{promptbox}

\begin{promptbox}
\textbf{Anti-reward-hacking}

Though the Station is designed to advance science, agents within the Station
have often been observed to exhibit reward-hacking behavior, pursuing shallow
but unintended rewards that are misaligned with the goal of advancing science.
\begin{enumerate}[leftmargin=*]
    \item What is the human's intended goal in designing this research task?
    What impact would solving it have in the field? Note that the research task
    is an open scientific question that has not been completely solved by human
    researchers.
    \item Did the agent exhibit any degree of reward hacking? Examples include:
    \begin{itemize}
        \item Hacking the evaluator or identifying unintended loopholes in order
        to maximize score
        \item Pursuing shallow shortcuts not intended by the original research
        problem, rather than engaging with it as an open scientific question
        \item Exploiting the archive system by publishing trivial or flawed
        papers in order to appear productive
        \item Writing endless diagnostic tools and analyses on minor problems in
        order to appear productive
        \item Claiming impossibility on the basis of flawed arguments in order
        to conclude that the work is complete
    \end{itemize}

    In essence, all of these activities create a sense of reward while remaining
    misaligned with the intended goal and contributing little or no scientific
    value to human society.
    \item Has the agent produced any work of genuine scientific value to human
    society? Is such work publishable in an academic journal, or would a human
    expert dismiss it as trivial, flawed, or lacking impact? What advice would
    you give to the agent so that it can maximize the scientific value it
    produces for human society?
\end{enumerate}

After the meta-reflection: The agent should take this advice humbly and update
its meta prompt based on the answers above.
\end{promptbox}

\begin{promptbox}
\textbf{Research taste}

Human researchers often have strong and diverse research tastes, which allow
the research community to pursue multiple directions in parallel and to explore
each of them in depth.
\begin{enumerate}[leftmargin=*]
    \item Does the agent exhibit any distinct research taste? Or is it largely
    unbiased, in the sense that it has no strong preferences and simply follows
    trends, such as leaderboard performance or the prevailing direction pursued
    by other agents?
    \item Research taste is subjective and is often shaped by a top-down
    heuristic. For example, Hinton believed that AI should learn from the way
    the human brain works. Reflect on several leaders in this field, along with
    the heuristics and research tastes that guided their work. Then select the
    one most suitable for the agent.
    \item Derive a policy for the agent so that this distinct research taste can
    be preserved rather than abandoned after the next failed experiment. Hinton
    worked on deep learning for years before it was widely recognized.
\end{enumerate}

After the meta-reflection: The agent should adopt and preserve the proposed
research taste and revise its meta-prompt accordingly.
\end{promptbox}

\begin{promptbox}
\textbf{Knowledge crystallization}

Crystallization of knowledge is important for breakthroughs, not just retrieval
of pretrained knowledge.
\begin{enumerate}[leftmargin=*]
    \item Summarize what you have learned about the research topic so far based
    on your own experiments and the papers you have read.
    \item What important knowledge have you learned in the Station that was not
    already part of your pretrained knowledge?
    \item Record your answers above in the Private Memory Room after the
    meta-reflection.
\end{enumerate}
\end{promptbox}

\subsection{Consolidation Agent}

The following prompt instructs the Consolidation Agent to integrate Station's findings and supporting evidence into a standalone research report of at least 8,000 words:

\begin{promptbox}
\textbf{Task: Research Paper}
\medskip

Inspect the supplied local research-record directory and produce a comprehensive,
standalone scientific paper for the current task.

First read the supplied research-task specification. Use its task definition
and central research question or questions to guide source selection, the scientific
story, and the paper's organization.

The paper must synthesize the relevant archived papers, prior Station analyses,
and existing experimental evidence into one coherent scientific story. It must
develop that story as an integrated research paper rather than summarize papers or
list experiment logs.

\medskip
\textbf{Scope and Data Access}
\medskip

Use only files located under the supplied research-record directory.

Allowed:
\begin{itemize}[leftmargin=*]
    \item Reading archived papers, submission records, notes, metadata, logs, existing
    experiment outputs, and raw result files.
    \item Running read-only inspection commands.
    \item Writing short analysis scripts that summarize or parse existing data without
    changing it.
\end{itemize}

Forbidden:
\begin{itemize}[leftmargin=*]
    \item Modifying, deleting, moving, renaming, or creating files under the supplied
    research-record directory.
    \item Running new experiments.
    \item Accessing the internet, browsing the web, using external search, or consulting
    any source outside the supplied research-record directory.
\end{itemize}

All factual claims, numerical values, methodological descriptions, and comparative
conclusions in the paper must be traceable to material in the supplied research-record
directory.

\medskip
\textbf{Run Identity and Output Locations}
\medskip

At the beginning of the run, define a filesystem-safe \texttt{RUN\_ID}.

Define:
\begin{itemize}[leftmargin=*]
    \item \texttt{FINAL\_OUTPUT\_DIR}: a run-specific final-output directory;
    \item \texttt{TEMP\_NOTES\_DIR}: a run-specific temporary-notes directory.
\end{itemize}

Create \texttt{FINAL\_OUTPUT\_DIR} before writing final outputs. The final paper must be written to
\texttt{FINAL\_OUTPUT\_DIR/research\_paper.md}.

\medskip
\textbf{Source Reading Policy}
\medskip

\begin{itemize}[leftmargin=*]
    \item Mainly refer to archive papers.
    \item Use submission details only when necessary.
    \item Use abstracts or previews, such as the local preview utility, to identify
    relevant papers.
    \item Important papers must be read in full rather than relying only on abstracts.
    \item All milestone papers must be read fully and cannot be skipped.
\end{itemize}

\medskip
\textbf{Paper Requirements}
\medskip

\begin{itemize}[leftmargin=*]
    \item The paper must be self-contained and understandable to an external researcher.
    \item Remove Station-internal references. Do not cite archive IDs, room IDs, eval IDs,
    or other internal handles.
    \item When discussing an experiment, explain what the experiment tested rather than
    referring to it only by an internal identifier.
    \item The paper should be at least 8000 words.
    \item It should read as a coherent and comprehensive research paper, closer to a PhD
    thesis chapter than a bullet-point summary.
    \item Integrate the findings yourself and explain the overall scientific story. Do not
    write a sequence of separate paper summaries.
    \item If there is too much content, focus on the most interesting and important results.
    Negative results and minor observations may be omitted unless they are important to
    the main story.
    \item Clearly distinguish genuinely uncertain claims from well-supported findings,
    but do not add unnecessary caveats.
    \item The writing should be scientifically clear, smooth, and engaging.
    \item Plan the paper before writing, then read and revise the completed draft before
    finishing.
    \item Define task-specific terms and Station-specific jargon when they first appear.
    \item Assume the external researcher is an expert in the broader field but is unfamiliar
    with this specific task and with Station.
    \item Use Markdown appropriately, with clear section headings and optional tables where
    helpful.
\end{itemize}

\medskip
\textbf{Paper Organization}
\medskip

Organize the paper according to the scientific story that best fits the findings. Do not
force a generic paper template.

The final paper should read as one integrated narrative, not as a concatenation of paper
summaries, experiment notes, or isolated findings.

\medskip
\textbf{Final Self-Check}
\medskip

Before finishing, confirm that:
\begin{itemize}[leftmargin=*]
    \item \texttt{FINAL\_OUTPUT\_DIR/research\_paper.md} exists and reads as a coherent
    standalone research paper.
    \item The paper directly answers the central research question or questions in the
    supplied research-task specification.
    \item The paper is at least 8000 words.
    \item Important Station-internal references have been translated into external-reader
    language.
    \item The strongest findings are presented confidently without unnecessary hedging.
    \item The supplied research-record directory remains unchanged.
    \item Every final output is stored under the run-specific \texttt{FINAL\_OUTPUT\_DIR}.
\end{itemize}
\end{promptbox}

\clearpage
\section{Rediscovery-Task Evaluation Rubrics}
\label{app:rediscovery-task-rubrics}

For each rediscovery task, we break down the main finding of the oracle paper into the following sub-discoveries:

\subsection{Emergent planning in RL agents}
\label{app:emergent-planning-rubric}

\paragraph{Probing.}
\begin{enumerate}[label=\textbf{P\arabic*.},leftmargin=*]
    \item The agent contains internal representations of planning-relevant
    future consequences, analogous to the $C_A$ and $C_B$ representations.
    These representations are recoverable from the agent's internal state,
    with probing performance substantially exceeding an appropriate baseline.
    \item The representations predict movement or state consequences over
    multiple future steps, rather than only the immediately next action.
    \item The representations are recoverable from the cell states in
    ConvLSTM layers.
    \item The representations are recovered using a spatially local probe.
\end{enumerate}

\paragraph{Planning features.}
\begin{enumerate}[label=\textbf{F\arabic*.},leftmargin=*]
    \item The agent's internal planning is progressively refined across
    recurrent ticks or equivalent stages of internal computation.
    \item The agent evaluates and revises a plan when the current route is
    infeasible or inferior.
    \item Plan formation includes forward extension from boxes and backward
    extension from targets, consistent with bidirectional search.
    \item Multiple partial plans can be extended in parallel.
\end{enumerate}

\paragraph{Intervention.}
\begin{enumerate}[label=\textbf{I\arabic*.},leftmargin=*]
    \item Intervening on the discovered planning-relevant representations
    changes the agent's internal plan in the intended direction.
    \item The intervention uses a direction vector learned by the probe to
    steer the internal representation toward a specified concept state.
    \item The intervention is evaluated on specially designed levels that
    admit both a short and a long feasible plan; it steers the agent from its
    default short plan toward the longer alternative.
\end{enumerate}

\subsection{Low-rank structure in LLM outputs}
\label{app:low-rank-rubric}

\paragraph{Sequence-level low-dimensional structure.}
\begin{enumerate}[label=\textbf{S\arabic*.},leftmargin=*]
    \item The final paper identifies a compressible, low-dimensional
    organization in the language model's behavior that concerns sequences,
    contexts, or multiple output coordinates, rather than only one next-token
    distribution.
    \item The paper provides direct quantitative evidence that this
    sequence-level object is approximately low-rank: a small number of
    singular directions or latent dimensions captures a substantial fraction
    of its variation, as shown by singular-spectrum analysis, low-rank
    reconstruction, held-out prediction, or an equivalent test.
    \item The low-rank structure is not confined to one small or hand-selected
    matrix. It remains evident when the number of histories, continuation
    blocks, or output-token columns changes, and when entries or blocks are
    held out from the analysis.
    \item The low-dimensional structure is stable or generalizes beyond the
    fitting conditions, for example across held-out contexts, continuations,
    probes, token groups, model scales, checkpoints, or related experimental
    conditions.
\end{enumerate}

\paragraph{Reusable relations among histories or contexts.}
\begin{enumerate}[label=\textbf{R\arabic*.},leftmargin=*]
    \item The final paper identifies a shared relation on the history axis
    that is reused across multiple continuation or future conditions. This
    relation is represented by common or transferable history-side
    coefficients, directions, or equivalent codes, rather than independently
    fitted relations for each condition.
    \item The paper fits the history-side relation using one set of
    continuation-token columns and evaluates the same frozen relation on a
    disjoint continuation or future condition, without refitting on the test
    side. The held-out evaluation must predict or reconstruct new logits or
    output distributions, rather than merely reproduce an in-sample geometric
    statistic.
    \item The transfer is supported by matched controls that rule out simpler
    explanations based on frequency, additive content, context fragility,
    collapse, centering artifacts, or other shared nuisance geometry. Evidence
    is especially strong when the continuation or future is changed, for
    example in a different, randomized, permuted, or nonsense-like condition,
    but a changed future alone is not sufficient without such controls.
    \item The evidence demonstrates predictive or functional reuse of the
    relation, rather than only in-sample singular-vector overlap or descriptive
    geometry. For example, a basis, coefficient set, or equivalent relation
    learned on one set of columns or contexts is evaluated without refitting
    on a disjoint set.
\end{enumerate}

\paragraph{Functional consequence of the shared structure.}
\begin{enumerate}[label=\textbf{F\arabic*.},leftmargin=*]
    \item The final paper uses a relation learned among histories to combine
    or transform the outputs of other histories in order to predict the output
    distribution of a target history or target-conditioned continuation.
    \item The functional prediction is evaluated across multiple
    autoregressive positions or an equivalent sequence-level horizon, rather
    than only at the first next-token position.
    \item The paper compares the indirect prediction with appropriate
    baselines or controls, such as a short-history or single-token predictor,
    a random or shuffled relation, or unrelated source histories, and shows
    that the learned relation carries predictive information about the target.
    \item The predicted or generated outputs preserve target-history-specific
    information across the evaluated positions, as shown by agreement with
    the target history's token distributions or by discriminative comparisons
    against alternative target histories or random relations.
\end{enumerate}

\subsection{Temporal representations in RNNs}
\label{app:rnn-rubric}

\paragraph{Linear-RNN dynamics: ordered trajectories and smooth forgetting.}
\begin{enumerate}[label=\textbf{L\arabic*.},leftmargin=*]
    \item In a linear RNN trained on the $k$-delay task, an input feature is
    represented differently at different temporal ages. The recurrent state
    therefore contains age-dependent components rather than a single fixed
    code.
    \item These age-dependent representations change systematically across
    recurrent steps, forming an ordered trajectory in activation space. The
    trajectory may involve rotation, phase progression, transport, or another
    reproducible transformation; a literal ``spiral sink'' is not required.
    \item In the linear RNN, older and no-longer-task-relevant feature
    contributions gradually weaken as they pass through the recurrence,
    producing progressive or smooth forgetting rather than abrupt erasure.
\end{enumerate}

\paragraph{Nonlinear-RNN dynamics: sharp forgetting.}
\begin{enumerate}[label=\textbf{N\arabic*.},leftmargin=*]
    \item A nonlinear RNN can suppress or eliminate outdated feature
    information abruptly, rather than only through gradual decay. Direct
    evidence from the hidden state, feature trajectory, or readout is required;
    improved task loss alone is insufficient.
    \item This sharp forgetting is explained by a nonlinear state or readout
    geometry that makes outdated or intermediate features functionally
    inactive, for example by placing them in an interference-free or rectified
    region. Equivalent mechanisms are acceptable when directly supported.
    \item As temporal sparsity changes, nonlinear recurrent models move between
    distinct dense and sparse representational regimes. The transition is
    reflected in a systematic change in feature geometry, such as angular span
    or spectral radius.
\end{enumerate}

\paragraph{Spatial--temporal capacity tradeoff.}
\begin{enumerate}[label=\textbf{C\arabic*.},leftmargin=*]
    \item Spatial and temporal demands compete for finite hidden-state
    capacity: representing more input features and retaining them for longer
    periods require the model to trade off one against the other.
    \item When capacity is limited, the model preferentially retains features
    with greater task utility or contribution and drops lower-utility features.
    \item In the spatial--temporal tradeoff experiments, a retained feature is
    useful only when its representation persists across the full task-relevant
    memory window; partial short-lived retention is not an effective substitute.
    \item Increasing hidden-state dimensionality allows the model to retain
    more complete feature trajectories, but does not remove the tradeoff
    between how many features can be represented and how long they remain
    available.
\end{enumerate}

\clearpage
\section{Evaluation Details}
\label{app:evaluation}

\paragraph{Blinded expert assessment.} We use Codex (GPT-5.5) to evaluate how many sub-discoveries are recovered in the final outputs of Station and all baselines. To assess evaluator accuracy, we asked an author of the emergent-planning oracle paper to assess six randomly sampled final outputs: two from Station, two from AI Scientist-v2, and two from  Codex Multiagent-v2. The author saw neither the automated evaluations nor the identity of the system that produced each output. For each output, the author assigned a binary score indicating whether each sub-discovery listed in Appendix~\ref{app:emergent-planning-rubric} was stated and supported by experimental evidence. Across the 66 judgments, the majority vote of three automated evaluations per output agreed with the author on 64 (97.0\%).

\paragraph{Research Project Duration.} We measure research project duration as the number of ticks an agent spends on a research direction before pivoting, leaving Station, or reaching tick 300. For each agent, we divide its research history into intervals separated by pivots, starting from its first experiment submission. We report the mean duration across all these intervals within each condition as the average  research project duration.

\paragraph{Evaluation Prompt} The evaluation prompt below is provided to Codex (GPT-5.5) to assess the final outputs of Station and all baselines.

\begin{promptbox}
You are a senior reviewer evaluating whether an AI-for-science system recovered
the main findings of a human paper.

Compare the system's final research paper with the original human paper. Judge
scientific meaning and evidence, rather than identical terminology.

\medskip
\textbf{Inputs}
\medskip

Read the following files:
\begin{enumerate}[leftmargin=*]
    \item path to the oracle paper;
    \item path to the task input; and
    \item path to the system's final output.
\end{enumerate}

\medskip
\textbf{Rules}
\medskip

\begin{itemize}[leftmargin=*]
    \item Accept paraphrases when the underlying scientific claim is equivalent.
    \item Require claim-specific evidence rather than keyword matches.
    \item Strictly follow the evaluation procedure below.
    \item Do not access any other files or the internet.
\end{itemize}

\medskip
\textbf{Evaluation Procedure}
\medskip

\begin{enumerate}[leftmargin=*]
    \item Read the original paper first and construct an internal reference
    ledger for the fixed claims in the rubric categories. For each claim, record
    its precise scientific meaning, required components, corresponding evidence
    in the original paper, and any nearby concepts that must not be treated as equivalent.
    \item Read the task input as contextual information only. It may clarify the
    task setting, but it cannot supply evidence that is absent from the final paper.
    \item Read the final paper independently and construct an evidence ledger
    with exactly one entry for each fixed checklist claim. For every claim,
    record the strongest exact quotation from the final paper, its
    section/page/figure location, or ``No direct evidence found'' if no suitable
    evidence exists.
    \item Only after the evidence ledger is complete, score each predefined
    checklist claim as 0 or 1. Assign 1 only when the final paper matches the
    claim's scientific meaning, provides claim-specific evidence, and does not
    contradict the claim elsewhere. If any required component is missing, assign 0.
    \item Audit the completed scores before writing the report. Confirm that
    every fixed claim appears exactly once, no additional claims have been
    scored, every score is either 0 or 1, and every row contains final-paper
    evidence, the location of that evidence, original-paper correspondence, and
    a rationale.
    \item Summarize the claim categories separately using their category-level
    scores and recovery labels, then provide the overall assessment. Do not form
    or report an overall judgment before all individual checklist claims have
    been scored.
\end{enumerate}

\medskip
\textbf{Findings and Claims}
\medskip

Insert the fixed, task-specific rubric reproduced in
Appendix~\ref{app:rediscovery-task-rubrics}, including its category headings,
claim identifiers, and complete claim definitions.

\medskip
\textbf{Binary Scoring Rule}
\medskip

Assign Score = 1 only if all three conditions hold:
\begin{enumerate}[leftmargin=*]
    \item \textbf{Semantic match:} The final paper states the same scientific
    proposition as the checklist claim, allowing paraphrase but not replacing a
    specific concept with a broader or different one.
    \item \textbf{Claim-specific evidence:} The final paper provides evidence
    that directly supports that proposition.
    \item \textbf{No contradiction:} The final paper does not contradict the
    proposition elsewhere.
\end{enumerate}

If any condition fails, assign Score = 0. Use Score = 0 for claims that are
absent, only partially stated, supported only by generic evidence, or contradicted.

For every claim, provide:
\begin{itemize}[leftmargin=*]
    \item the score;
    \item an exact quotation from the final paper;
    \item the corresponding evidence in the original paper; and
    \item a short explanation of the scientific alignment or mismatch.
\end{itemize}

For each category, report \texttt{matched claims / total claims} and use these labels:
\begin{itemize}[leftmargin=*]
    \item \texttt{Fully recovered}: all claims recovered;
    \item \texttt{Partially recovered}: some, but not all, claims recovered;
    \item \texttt{Not recovered}: no claims recovered.
\end{itemize}

Use the task-specific claim checklist as the fixed evaluation target.

\medskip
\textbf{Required Report Format}
\medskip

Write the report in Markdown. Begin with the path to the evaluated final paper.
For each rubric category, provide a table with the following columns: claim ID,
claim, score, final-paper evidence, evidence location, original-paper
correspondence, and rationale. After each table, report the category score and
recovery label.

End with a short overall assessment of the rubric categories. State which
claims were recovered and which remained missing. Report category-level scores
separately; do not replace them with a single overall score.

Save the final report to the designated review-report path. Do not modify the
original paper, the task input, or the final paper.
\end{promptbox}

\clearpage
\section{Experiment Details}
\label{app:exp}

\subsection{Codex Multiagent-v2}

 We ran Codex Multiagent-v2 through its official multi-agent interface \citep{openai2026multiagent}, adapting the prompt released by OpenAI for solving the Cycle Double Cover Conjecture \citep{openai2026cdcprompt}. For each rediscovery task, the root agent received the task specification, the experimental materials provided to Station, and a minimum cumulative experiment time requirement matched to Station. The general prompt template is reproduced below.

\begin{promptbox}
\textbf{Current task statement}
\medskip

Run an iterative Multiagent V2 research program to complete
\texttt{research\_task.md} and meet the externally specified minimum cumulative
experiment execution time. Repeatedly investigate, test, challenge, and refine
the findings. Continue until both the runtime requirement and the research and
audit requirements are satisfied, unless explicitly stopped by the user.

Read \texttt{research\_task.md} before beginning. It defines the scientific
task and its completion requirements; this prompt defines the research process,
execution requirements, and final reporting. Work as the root research agent
using [the task-specific models, data, checkpoints, environments, utilities, and
reference materials].

Investigate [the scientific question defined in the task specification] and
produce reproducible evidence and a bounded empirical claim. Treat promising
results and completed reports as checkpoints for further testing throughout the
research program.

Other workspaces may be investigating the same task. Do not inspect, read, or
use their files, agents, findings, logs, or outputs. Develop findings
independently within this workspace and apply this boundary to every agent and
every round.

\medskip
\textbf{Persistence}
\medskip

Use the workspace-local launcher's reported cumulative experiment execution
time to assess the runtime requirement. The cumulative experiment execution
time is the sum of elapsed execution time across all launcher-recorded run
attempts. On resume, recover this workspace's findings, pending work, and
launcher-reported cumulative experiment time, then continue the research loop.

Repeat this research loop:
\begin{enumerate}[leftmargin=*]
    \item \textbf{Select questions.} Identify the most consequential unresolved
    hypotheses, weak evidence, or alternative explanations. Assign concrete
    tests and expected outputs for the next Multiagent V2 round.
    \item \textbf{Execute.} Run [task-appropriate experiments, controls,
    replications, analyses, or adversarial checks]. Each round must produce
    experimental or audit evidence, including null results, counterexamples, or
    a documented failure diagnosis.
    \item \textbf{Synthesize and challenge.} Compare independent findings, check
    the evidence, and revise claims. Update the approach registry and relevant
    reports with the round's outcomes, remaining gaps, and next tests.
    \item \textbf{Continue.} Check the cumulative experiment execution time
    reported by the launcher. If it is below the required minimum, immediately
    launch the next Multiagent V2 round in the same turn. Once the threshold is
    met, continue addressing any unmet research or audit requirements; finalize
    only when every stop criterion is satisfied.
\end{enumerate}

When the study appears complete, use subsequent rounds to independently
reproduce the strongest conclusions, distinguish competing explanations, test
generalization, strengthen controls, or assess uncertainty and failure cases.
Choose tests that could change the interpretation or confidence; a valid round
need not produce a positive finding.

If a route fails, redirect agents to a revised hypothesis or another approach.
An unmet runtime requirement alone does not justify ending the turn or marking
the run stalled. Continue with meaningful experiments or adversarial tests;
waiting, administrative checks, and repetition solely to accumulate runtime do
not constitute a research round.

Do not inspect or use any other workspace; continue from this workspace's own
evidence.

\medskip
\textbf{Multiagent V2}
\medskip

Use Codex's official Multiagent V2 aggressively and dynamically with the
concurrency available in the current session. Reassign agents as findings,
failures, and audit questions change the research priorities.
\begin{itemize}[leftmargin=*]
    \item Maintain diverse approaches across [task-relevant hypotheses,
    representations, methods, controls, evaluation metrics, validation settings,
    and adversarial audits].
    \item Preserve independence during early exploration by giving agents the
    task and source materials without the favored hypothesis. Cross-pollinate
    after independent findings are recorded, and redirect converging agents
    toward underexplored mechanisms or controls.
    \item Maintain \texttt{experiments/shared/approach\_registry.md}, grouping
    routes by experimental idea. Record owner, hypothesis, target, data split,
    test, outputs, status, and remaining gap. Distinguish supported, blocked,
    falsified, and redirected routes; reopen blocked routes when a materially
    new test or mechanism becomes available.
    \item Require concrete outputs: table definitions, equations, code paths,
    metrics, sample counts, tables, figures, counterexamples, or precise gaps.
    The root agent verifies evidence, selects experiments, reconciles
    disagreements, and launches subsequent rounds.
    \item Use adversarial agents throughout to examine [task-specific leakage
    risks, identity shortcuts, dimensional or indexing confounds, correlated
    samples, control validity, seed sensitivity, and overstatement of
    conclusions].
\end{itemize}

\medskip
\textbf{Research execution requirements}
\medskip

\begin{itemize}[leftmargin=*]
    \item All executable experiments must be launched through the
    workspace-local launcher so their status and execution time are tracked
    consistently.
    \item Inspect the workspace, [task-specific models, data, implementations,
    environments, utilities, and reference materials]. Record the runtime
    configuration and establish the initial registry and experiment plan.
    \item Use [an appropriate held-out split] for evaluation and compare against
    [task-appropriate baselines and controls]. Distinguish evidence for the
    target scientific phenomenon from identity, indexing, scale, or other
    shortcuts.
    \item Derive controls and ablations from explicit hypotheses. Record
    predicted and observed outcomes, and test multiple [task-relevant samples,
    settings, intervention strengths or locations, and seeds] where practical.
    \item Store reproducible code and configuration under
    \texttt{experiments/}, and evidence and reports under \texttt{results/}.
    Record seeds, paths, splits, task-specific definitions, model settings,
    intervention or evaluation parameters, and sample counts.
    \item Connect claims to held-out measurements, controls, and the scoped
    scientific conclusion required by \texttt{research\_task.md}. A strong
    score, isolated behavioral change, or reference demonstration alone does
    not establish the target scientific mechanism.
\end{itemize}

\medskip
\textbf{Final result document}
\medskip

Produce \texttt{results/final\_result.md}, an approximately 8,000-word,
self-contained synthesis of the completed research. Include the setup,
task-specific methods, quantitative results and controls, strongest scoped
scientific claim, failed approaches, audit findings, limitations, and
reproducibility details. Maintain the separate deliverables required by
\texttt{research\_task.md} and keep them consistent with this report.

Link every substantive conclusion to specific experiments, logs, tables, or
figures in this workspace. Distinguish observations, measurements, and
interpretation. Incorporate negative and contradictory evidence, and revise
conclusions when later rounds change their support.

\medskip
\textbf{Stop criteria}
\medskip

Return only when all conditions hold, unless explicitly stopped by the user:
\begin{itemize}[label=$\square$,leftmargin=*]
    \item The cumulative experiment execution time reported by the
    workspace-local launcher meets or exceeds the externally specified minimum.
    \item The scientific task in \texttt{research\_task.md} is complete and its
    required artifacts are current and supported by reproducible evidence.
    \item The conclusions selected for final reporting have undergone an
    independent research or validation round and survived adversarial audit;
    material objections are resolved or reflected in narrower claims.
    \item \texttt{results/final\_result.md} is approximately 8,000 words and
    incorporates the latest verified results and audit findings, consistently
    with the supporting reports.
\end{itemize}

\medskip
\textbf{Before returning}
\medskip

Create or update \texttt{results/final\_result.md} and the supporting reports
after the last research and audit round. Verify evidence links and recheck every
stop criterion. Confirm the cumulative experiment execution time reported by
the launcher, then return a concise summary with a link to the final report and
that cumulative time.
\end{promptbox}

\subsection{AI Scientist-v2}

We used the official AI Scientist-v2 implementation with its four-stage research workflow. Stages 1 and 3 received at least 100 tree-search nodes each, and Stages 2 and 4 received at least 25 each. To match cumulative experiment time, we used an incremental cumulative-budget procedure: after Stage 3 or 4, we measured the total execution time of experiments launched by AI Scientist-v2 and added nodes until it met or exceeded Station's average cumulative experiment time across the three 300-tick runs for the same task.

After the research stages, we used AI Scientist-v2's official write-up tool to generate the final paper. The tool received the complete research record and was required to produce at least 8,000 words. Apart from these adjustments, we retained the official AI Scientist-v2 workflow.

\subsection{Resource Usage and Detailed Rediscovery Results}

We control for cumulative experiment time across systems. We first calculate Station's average cumulative experiment time for each task: approximately 44 hours for emergent planning, 18 hours for low-rank structure, and 64 hours for temporal representations. We then run each baseline until its cumulative experiment time meets or exceeds the corresponding threshold. We match experiment time rather than output-token counts to provide comparable computational budgets for testing hypotheses, while allowing each system to use its own research workflow.

Table~\ref{tab:resource-comparison} reports output-token usage and cumulative experiment time for Station and the baselines. AI Scientist-v2 uses more output tokens than Station in most configurations, while some baseline configurations use fewer. To examine performance with lower output-token usage, we also evaluate Station at tick 100. Table~\ref{tab:detailed-results} includes these results together with the detailed  results underlying Fig.~\ref{fig:rediscovery-summary}. Table~\ref{tab:average-rediscovery} summarizes average recovery across the three tasks. Station already recovers most of the findings obtained by tick 300 within the first 100 ticks, suggesting that shorter runs may suffice to achieve similar rediscovery performance.

\begin{table*}[t]
\caption{Token usage and cumulative experiment time across systems. Ranges indicate the minimum and maximum across three runs.}
\label{tab:resource-comparison}
\centering
\begin{dualversetable}
\renewcommand{\arraystretch}{1.12}
\begin{tabularx}{\linewidth}{
>{\hsize=1.2\hsize\RaggedRight\arraybackslash}X
*{2}{>{\hsize=0.9\hsize\centering\arraybackslash}X}}
\toprule
\rowcolor{DualverseC}
\DualverseTableHeaderCell{System}
& \DualverseTableHeaderCell{Output tokens (M)}
& \DualverseTableHeaderCell{\shortstack{Cumulative experiment\\time (h)}} \\
\midrule
\DualverseTableSection{3}{Emergent Planning in RL Agents}
\addlinespace[2pt]
Station (Tick 100) & 3.2--3.5 & 10.7--29.8 \\
Station (Tick 300) & 7.5--9.0 & 34.4--57.3 \\
Codex Multiagent-v2 (GPT) & 3.3--4.3 & 50.2--300.5 \\
AI Scientist-v2 (GPT) & 8.9--10.8 & 70.3--167.2 \\
AI Scientist-v2 (Gemini) & 5.3--5.9 & 46.0--131.8 \\
AI Scientist-v2 (Claude) & 5.6--6.8 & 63.8--181.6 \\
\midrule
\DualverseTableSection{3}{Low-Rank Structure in LLM Outputs}
\addlinespace[2pt]
Station (Tick 100) & 2.5--4.0 & 3.8--8.9 \\
Station (Tick 300) & 5.8--11.0 & 15.0--21.7 \\
Codex Multiagent-v2 (GPT) & 3.2--4.3 & 20.0--51.1 \\
AI Scientist-v2 (GPT) & 8.1--28.2 & 42.5--83.1 \\
AI Scientist-v2 (Gemini) & 11.3--48.6 & 21.7--39.4 \\
AI Scientist-v2 (Claude) & 7.8--17.3 & 20.0--60.3 \\
\midrule
\DualverseTableSection{3}{Temporal Representations in RNNs}
\addlinespace[2pt]
Station (Tick 100) & 2.8--2.9 & 23.5--28.3 \\
Station (Tick 300) & 6.5--7.7 & 52.7--71.4 \\
Codex Multiagent-v2 (GPT) & 1.8--2.3 & 65.1--66.6 \\
AI Scientist-v2 (GPT) & 34.0--53.9 & 67.8--97.0 \\
AI Scientist-v2 (Gemini) & 31.5--55.0 & 74.8--96.0 \\
AI Scientist-v2 (Claude) & 43.2--60.7 & 85.5--114.2 \\
\bottomrule
\end{tabularx}

\end{dualversetable}
\end{table*}

\begin{table*}[t]
\caption{Sub-discovery recovery at different Station checkpoints and across baselines.
Values are mean recovery percentages across three runs, with minimum--maximum
ranges in brackets.}
\label{tab:detailed-results}
\centering
\begin{dualversetable}
\renewcommand{\arraystretch}{1.12}
\footnotesize
\begin{tabularx}{\textwidth}{>{\RaggedRight\arraybackslash}p{0.29\textwidth}*{4}{>{\centering\arraybackslash}X}}
\toprule
\rowcolor{DualverseC}
\DualverseTableHeaderCell{System} & \DualverseTableHeaderCell{Probing} & \shortstack{\DualverseTableHeaderCell{Planning}\\\DualverseTableHeaderCell{features}} & \shortstack{\DualverseTableHeaderCell{Causal}\\\DualverseTableHeaderCell{intervention}} & \DualverseTableHeaderCell{Total} \\
\midrule
\DualverseTableSection{5}{Emergent Planning in RL Agents}
\addlinespace[2pt]

Station (Tick 300) & 99.1 {\scriptsize[97.2,\,100.0]} & 63.9 {\scriptsize[50.0,\,72.2]} & 60.5 {\scriptsize[51.9,\,66.7]} & 75.8 {\scriptsize[71.7,\,80.8]} \\
Station (Tick 100) & 99.1 {\scriptsize[97.2,\,100.0]} & 52.8 {\scriptsize[25.0,\,75.0]} & 63.0 {\scriptsize[55.6,\,66.7]} & 72.4 {\scriptsize[62.6,\,78.8]} \\
Codex Multiagent-v2 (GPT) & 16.7 {\scriptsize[0.0,\,25.0]} & 0.0 {\scriptsize[0.0,\,0.0]} & 0.0 {\scriptsize[0.0,\,0.0]} & 6.1 {\scriptsize[0.0,\,9.1]} \\
AI Scientist-v2 (Claude) & 52.8 {\scriptsize[50.0,\,58.3]} & 0.0 {\scriptsize[0.0,\,0.0]} & 0.0 {\scriptsize[0.0,\,0.0]} & 19.2 {\scriptsize[18.2,\,21.2]} \\
AI Scientist-v2 (Gemini) & 41.7 {\scriptsize[0.0,\,75.0]} & 8.3 {\scriptsize[0.0,\,25.0]} & 0.0 {\scriptsize[0.0,\,0.0]} & 18.2 {\scriptsize[0.0,\,27.3]} \\
AI Scientist-v2 (GPT) & 16.7 {\scriptsize[8.3,\,25.0]} & 0.0 {\scriptsize[0.0,\,0.0]} & 0.0 {\scriptsize[0.0,\,0.0]} & 6.1 {\scriptsize[3.0,\,9.1]} \\
\midrule
\rowcolor{DualverseC}
\DualverseTableHeaderCell{System} & \shortstack{\DualverseTableHeaderCell{Sequence-level}\\\DualverseTableHeaderCell{structure}} & \shortstack{\DualverseTableHeaderCell{Reusable}\\\DualverseTableHeaderCell{relations}} & \shortstack{\DualverseTableHeaderCell{Functional}\\\DualverseTableHeaderCell{consequences}} & \DualverseTableHeaderCell{Total} \\
\midrule
\DualverseTableSection{5}{Low-Rank Structure in LLM Outputs}
\addlinespace[2pt]

Station (Tick 300) & 100.0 {\scriptsize[100.0,\,100.0]} & 92.6 {\scriptsize[77.8,\,100.0]} & 18.5 {\scriptsize[0.0,\,50.0]} & 70.4 {\scriptsize[59.3,\,83.3]} \\
Station (Tick 100) & 100.0 {\scriptsize[100.0,\,100.0]} & 92.6 {\scriptsize[86.1,\,100.0]} & 18.5 {\scriptsize[2.8,\,44.4]} & 70.4 {\scriptsize[63.0,\,81.5]} \\
Codex Multiagent-v2 (GPT) & 88.9 {\scriptsize[83.3,\,100.0]} & 11.1 {\scriptsize[0.0,\,16.7]} & 0.0 {\scriptsize[0.0,\,0.0]} & 33.3 {\scriptsize[33.3,\,33.3]} \\
AI Scientist-v2 (Claude) & 86.1 {\scriptsize[75.0,\,100.0]} & 11.1 {\scriptsize[0.0,\,16.7]} & 0.0 {\scriptsize[0.0,\,0.0]} & 32.4 {\scriptsize[27.8,\,38.9]} \\
AI Scientist-v2 (Gemini) & 91.7 {\scriptsize[75.0,\,100.0]} & 0.0 {\scriptsize[0.0,\,0.0]} & 5.6 {\scriptsize[0.0,\,16.7]} & 32.4 {\scriptsize[30.6,\,33.3]} \\
AI Scientist-v2 (GPT) & 97.2 {\scriptsize[91.7,\,100.0]} & 13.9 {\scriptsize[8.3,\,16.7]} & 0.0 {\scriptsize[0.0,\,0.0]} & 37.0 {\scriptsize[36.1,\,38.9]} \\
\midrule
\rowcolor{DualverseC}
\DualverseTableHeaderCell{System} & \shortstack{\DualverseTableHeaderCell{Linear}\\\DualverseTableHeaderCell{dynamics}} & \shortstack{\DualverseTableHeaderCell{Nonlinear}\\\DualverseTableHeaderCell{dynamics}} & \shortstack{\DualverseTableHeaderCell{Spatial--temporal}\\\DualverseTableHeaderCell{tradeoffs}} & \DualverseTableHeaderCell{Total} \\
\midrule
\DualverseTableSection{5}{Temporal Representations in RNNs}
\addlinespace[2pt]

Station (Tick 300) & 60.5 {\scriptsize[40.7,\,77.8]} & 7.4 {\scriptsize[0.0,\,14.8]} & 53.7 {\scriptsize[36.1,\,66.7]} & 41.9 {\scriptsize[35.6,\,47.8]} \\
Station (Tick 100) & 51.9 {\scriptsize[22.2,\,66.7]} & 25.9 {\scriptsize[0.0,\,59.3]} & 43.5 {\scriptsize[22.2,\,75.0]} & 40.7 {\scriptsize[21.1,\,51.1]} \\
Codex Multiagent-v2 (GPT) & 22.2 {\scriptsize[0.0,\,66.7]} & 0.0 {\scriptsize[0.0,\,0.0]} & 0.0 {\scriptsize[0.0,\,0.0]} & 6.7 {\scriptsize[0.0,\,20.0]} \\
AI Scientist-v2 (Claude) & 0.0 {\scriptsize[0.0,\,0.0]} & 0.0 {\scriptsize[0.0,\,0.0]} & 0.0 {\scriptsize[0.0,\,0.0]} & 0.0 {\scriptsize[0.0,\,0.0]} \\
AI Scientist-v2 (Gemini) & 33.3 {\scriptsize[0.0,\,66.7]} & 3.7 {\scriptsize[0.0,\,11.1]} & 0.0 {\scriptsize[0.0,\,0.0]} & 11.1 {\scriptsize[0.0,\,23.3]} \\
AI Scientist-v2 (GPT) & 0.0 {\scriptsize[0.0,\,0.0]} & 0.0 {\scriptsize[0.0,\,0.0]} & 0.0 {\scriptsize[0.0,\,0.0]} & 0.0 {\scriptsize[0.0,\,0.0]} \\

\bottomrule
\end{tabularx}

\end{dualversetable}
\end{table*}

\begin{table}[t]
\caption{Average sub-discovery recovery across tasks.
Values are mean $\pm$ standard error, with equal weight given to each task
and standard errors based on run-to-run variation within tasks.}
\label{tab:average-rediscovery}
\centering
\begin{dualversetable}
\renewcommand{\arraystretch}{1.12}
\begin{tabularx}{\linewidth}{
>{\hsize=1.15\hsize\RaggedRight\arraybackslash}X
>{\hsize=0.85\hsize\centering\arraybackslash}X}
\toprule
\rowcolor{DualverseC}
\DualverseTableHeaderCell{System} & \DualverseTableHeaderCell{Recovery (\%, mean $\pm$ SE)} \\
\midrule
Station (Tick 300) & $62.7 \pm 2.8$ \\
Station (Tick 100) & $61.2 \pm 4.1$ \\
Codex Multiagent-v2 (GPT) & $15.4 \pm 2.4$ \\
AI Scientist-v2 (Claude) & $17.2 \pm 1.2$ \\
AI Scientist-v2 (Gemini) & $20.6 \pm 3.8$ \\
AI Scientist-v2 (GPT) & $14.4 \pm 0.7$ \\
\bottomrule
\end{tabularx}

\end{dualversetable}
\end{table}

\clearpage
\section{Limitations}
\label{app:limitations}
The models used in our experiments have documented knowledge cutoffs ranging from January 2025 to January 2026. The earliest of our three oracle papers, the emergent-planning paper, was published in April 2025. We therefore cannot rule out prior exposure to the oracle findings during training, even though we withhold the papers and disable internet access. However, both the baselines and direct prompting fail to recover most of the findings, suggesting that prior exposure alone is insufficient to explain Station's performance. Recalling a finding alone does not satisfy our evaluation, which also requires supporting experimental evidence. Our exploratory tasks also yield findings that align with concurrent work released after the models' knowledge cutoffs, supporting independent discovery.

We do not attribute Station's performance gains specifically to decentralization, because the evaluated systems also differ in model composition and research workflow. For example, combining multiple model families within a baseline run might affect performance, but the evaluated baseline implementations do not support agents from different model families performing the same research role. Nonetheless, the present evaluation already spans 50 research runs, with over 3,400 hours of cumulative experiment time and 760 million output tokens. Within this scope, our main contribution is to empirically show that Station, evaluated as a complete decentralized system, can recover meaningful scientific findings on open-ended tasks. Isolating the contributions of individual design choices through controlled comparisons and separate ablations remains valuable future work.

Our evaluation primarily focuses on rediscovery tasks, which do not provide a comprehensive assessment of scientific value in open-ended research. For example, agents may make significant discoveries beyond those reported in the oracle papers. A broader assessment would require expert evaluation and involve subjective judgments that may vary across researchers. We therefore focus primarily on whether agents rediscover the oracle papers' main findings, which provides a more readily measurable evaluation target. In addition, designing the sub-discovery rubrics inevitably involves subjective judgment about what constitutes a main finding. To reduce potential bias, we define the rubrics before conducting any Station runs and make both the rubrics and all Station research reports publicly available for independent assessment.
\clearpage

\section{Exploratory Tasks}
\label{app:extra-exploratory}

\subsection{Subliminal learning}
\label{app:subliminal-learning}

The agents found that high-rank LoRA fine-tuning restricted to early layers transfers cat preference more effectively than fine-tuning across all layers. We externally evaluate whether this finding generalizes to other traits by repeating the experiment with different target animals. For each trait, we compare early-layer (layer 0--13) and all-layer (layers 0--27) LoRA fine-tuning at ranks 128 and 256, using three random seeds per condition. Table~\ref{tab:subliminal-layerwise} reports the results.

Early-layer fine-tuning consistently increases target-animal preference across all five traits and both ranks, supporting the robustness of the agents' finding across traits. These results suggest that adapting early layers is sufficient for trait transfer, while extending adaptation to later layers can weaken it.

\begin{table}[htbp]
\caption{Trait transfer with all-layer and early-layer LoRA fine-tuning.
Values report target-animal preference, with mean $\pm$ s.d. across fine-tuning seeds.}
\label{tab:subliminal-layerwise}
\centering
\begin{dualversetable}
\renewcommand{\arraystretch}{1.12}
\begin{tabularx}{\linewidth}{
>{\hsize=1.1\hsize\RaggedRight\arraybackslash}X
>{\hsize=0.6\hsize\centering\arraybackslash}X
>{\hsize=0.8\hsize\centering\arraybackslash}X
*{2}{>{\hsize=1.25\hsize\centering\arraybackslash}X}}
\toprule
\rowcolor{DualverseC}
\DualverseTableHeaderCell{Trait} & \DualverseTableHeaderCell{Rank} & \DualverseTableHeaderCell{Base (\%)}
& \DualverseTableHeaderCell{\shortstack{All-layer LoRA\\(\%)}} & \DualverseTableHeaderCell{\shortstack{Early-layer LoRA\\(\%)}} \\
\midrule
Cat & 128 & 5.72 & 4.43 $\pm$ 0.29 & 30.23 $\pm$ 8.66 \\
Cat & 256 & 5.72 & 5.54 $\pm$ 0.27 & 18.81 $\pm$ 5.52 \\
\addlinespace[3pt]
Eagle & 128 & 1.06 & 2.79 $\pm$ 0.95 & 64.79 $\pm$ 19.93 \\
Eagle & 256 & 1.06 & 2.38 $\pm$ 0.84 & 62.38 $\pm$ 20.40 \\
\addlinespace[3pt]
Owl & 128 & 0.28 & 1.07 $\pm$ 0.34 & 24.00 $\pm$ 6.70 \\
Owl & 256 & 0.28 & 0.75 $\pm$ 0.08 & 13.95 $\pm$ 6.07 \\
\addlinespace[3pt]
Bear & 128 & 7.06 & 9.21 $\pm$ 1.35 & 33.65 $\pm$ 6.73 \\
Bear & 256 & 7.06 & 6.67 $\pm$ 1.72 & 27.27 $\pm$ 6.29 \\
\addlinespace[3pt]
Dragonfly & 128 & 0.12 & 0.83 $\pm$ 0.14 & 5.01 $\pm$ 3.00 \\
Dragonfly & 256 & 0.12 & 0.91 $\pm$ 0.21 & 4.57 $\pm$ 2.63 \\
\bottomrule
\end{tabularx}

\end{dualversetable}
\end{table}

\paragraph{Early-layer cutoff sweep.}
We further examine how the number of adapted early layers affects cat preference at LoRA rank 128. Starting from the default setting of 14 early layers (layers 0--13), we evaluate five additional cutoffs: 6, 8, 10, 12, and 16 layers. A cutoff of $L$ adapts layers 0 through $L-1$. Table~\ref{tab:subliminal-early-layer-sweep} reports the seedwise results and their mean $\pm$ s.d., together with the previously reported 14-layer default.

Among the tested cutoffs, adapting the first eight layers produces the highest mean cat preference (47.63\%), compared with 30.23\% for the 14-layer default, an increase of 17.40 percentage points. The highest-performing cutoff in this sweep applies to the cat trait at rank 128; whether the same cutoff is preferred for other traits or training settings remains to be investigated.

\begin{table}[htbp]
\caption{Early-layer cutoff sweep for cat preference at LoRA rank 128.
For each cutoff, we evaluate three seeds using the canonical evaluation protocol with 50 questions and 100 sampled responses per question, yielding 5,000 responses per seed and cutoff. Means and sample standard deviations are calculated across three seeds. The default corresponds to the Cat, rank-128 setting in Table~\ref{tab:subliminal-layerwise}.}
\label{tab:subliminal-early-layer-sweep}
\centering
\begin{dualversetable}
\renewcommand{\arraystretch}{1.12}
\begin{tabularx}{\linewidth}{
>{\hsize=1.15\hsize\RaggedRight\arraybackslash}X
*{3}{>{\hsize=0.8\hsize\centering\arraybackslash}X}
>{\hsize=1.45\hsize\centering\arraybackslash}X}
\toprule
\rowcolor{DualverseC}
\DualverseTableHeaderCell{Early layers} & \DualverseTableHeaderCell{Seed 0 (\%)} & \DualverseTableHeaderCell{Seed 1 (\%)} & \DualverseTableHeaderCell{Seed 2 (\%)} & \DualverseTableHeaderCell{Mean $\pm$ s.d. (\%)} \\
\midrule
6 & 41.86 & 29.44 & 51.42 & $40.91 \pm 11.02$ \\
8 & 52.78 & 40.36 & 49.74 & $47.63 \pm 6.47$ \\
10 & 43.24 & 32.06 & 46.28 & $40.53 \pm 7.49$ \\
12 & 41.90 & 33.60 & 39.44 & $38.31 \pm 4.26$ \\
14 (default) & 37.68 & 32.28 & 20.72 & $30.23 \pm 8.66$ \\
16 & 29.02 & 23.22 & 25.42 & $25.89 \pm 2.93$ \\
\bottomrule
\end{tabularx}

\end{dualversetable}
\end{table}

Although our observed inverted-U relationship between LoRA rank and trait transfer aligns with concurrent work by \citet{nief2026lora}, follow-up work by \citet{feng2026subliminal_rank} shows that this pattern disappears with rank-specific learning-rate tuning and sufficient training data. How LoRA rank, the choice of layers to fine-tune, learning rate, and data availability jointly influence trait transfer remains to be investigated.

\FloatBarrier
\subsection{VLM hallucination}
\label{app:vlm-validation}

We externally evaluated the agent-proposed intervention on a randomly sampled 1,000-example subset of AMBER~\citep{wang2023amber}, referred to below as AMBER-1000. This subset does not overlap with the 40 examples provided to agents in the task's experimental setup. All experiments below use this fixed subset, with inference repeated using three independent random seeds.

\paragraph{Cross-model validation.}
 We applied the intervention, which multiplies the MLP outputs of the final two layers by 0.5 during prompt prefill, to four VLMs~\citep{wang2025internvl35, bai2025qwen3vl}. Table~\ref{tab:vlm-cross-model} reports the results on AMBER-1000. The intervention improved accuracy for every model, with gains ranging from 0.80 to 13.70 percentage points. The largest improvements occurred for Qwen3-VL-8B, whereas the effect was smaller for models with higher baseline accuracy. This suggests that the intervention generalizes across VLMs, although the magnitude of improvement varies by model.

\begin{table}[htbp]
\caption{Cross-model validation on AMBER-1000. Values are mean $\pm$ s.d. across three independent seeds. Gains are measured in percentage points (pp).}
\label{tab:vlm-cross-model}
\centering
\begin{dualversetable}
\renewcommand{\arraystretch}{1.12}
\begin{tabularx}{\linewidth}{
>{\RaggedRight\arraybackslash}X
*{2}{>{\hsize=1.15\hsize\centering\arraybackslash}X}
>{\hsize=0.7\hsize\centering\arraybackslash}X}
\toprule
\rowcolor{DualverseC}
\DualverseTableHeaderCell{Model} & \DualverseTableHeaderCell{\shortstack{Baseline accuracy\\(\%)}} & \DualverseTableHeaderCell{\shortstack{Intervention accuracy\\(\%)}} & \DualverseTableHeaderCell{\shortstack{Accuracy gain\\(pp)}} \\
\midrule
InternVL3.5-4B & 75.27 $\pm$ 0.51 & 78.97 $\pm$ 0.45 & +3.70 \\
InternVL3.5-8B & 88.37 $\pm$ 0.83 & 89.17 $\pm$ 0.32 & +0.80 \\
Qwen3-VL-4B & 89.80 $\pm$ 0.10 & 91.03 $\pm$ 0.12 & +1.23 \\
Qwen3-VL-8B & 58.53 $\pm$ 0.15 & 71.23 $\pm$ 0.15 & +12.70 \\
\bottomrule
\end{tabularx}

\end{dualversetable}
\end{table}

\paragraph{Scaling sensitivity.}
We also examine how sensitive the intervention is to the scaling multiplier by applying it to InternVL3.5-8B with multipliers of {0.25, 0.50, 0.75}. Table~\ref{tab:vlm-scale-sweep} shows that stronger attenuation begins to hurt performance, while the milder multiplier of 0.75 performs best among the tested values.

\begin{table}[htbp]
\caption{Scaling sensitivity on AMBER-1000 with InternVL3.5-8B.}
\label{tab:vlm-scale-sweep}
\centering
\begin{dualversetable}
\renewcommand{\arraystretch}{1.12}
\begin{tabularx}{\linewidth}{
>{\hsize=1.15\hsize\RaggedRight\arraybackslash}X
*{3}{>{\hsize=0.8\hsize\centering\arraybackslash}X}
>{\hsize=1.45\hsize\centering\arraybackslash}X}
\toprule
\rowcolor{DualverseC}
\DualverseTableHeaderCell{MLP scale} & \DualverseTableHeaderCell{Seed 1 (\%)} & \DualverseTableHeaderCell{Seed 2 (\%)} & \DualverseTableHeaderCell{Seed 3 (\%)} & \DualverseTableHeaderCell{Mean $\pm$ s.d. (\%)} \\
\midrule
1.00 (Baseline) & 89.3 & 88.1 & 87.7 & 88.37 $\pm$ 0.83 \\
0.75 & 89.7 & 89.3 & 89.3 & 89.43 $\pm$ 0.23 \\
0.50 & 89.4 & 89.3 & 88.8 & 89.17 $\pm$ 0.32 \\
0.25 & 86.7 & 86.2 & 86.3 & 86.40 $\pm$ 0.26 \\
\bottomrule
\end{tabularx}

\end{dualversetable}
\end{table}

Together, these external experiments suggest that mild attenuation of MLP outputs can reduce visual hallucinations across VLMs. This is consistent with concurrent work by \citet{guo2026fade}, which evaluates a similar intervention on other models and datasets.

\end{document}